%% file: main.tex
\documentclass[twoside]{article}

\usepackage[accepted]{aistats2023}
\usepackage[round]{natbib}

\usepackage{microtype}
\usepackage{graphicx}
\usepackage{subfigure}
\usepackage{multirow}
\usepackage[utf8]{inputenc} 
\usepackage[T1]{fontenc}    
\usepackage{hyperref}       
\usepackage{url}            
\usepackage{booktabs}       
\usepackage{amsfonts}       
\usepackage{nicefrac}       
\usepackage{xcolor}         
\usepackage{IEEEtrantools}

\input{commands}

\begin{document}

%

%

\runningtitle{A Sheaf-Theoretic Framework for Knowledge Graph Embedding}
\twocolumn[

\aistatstitle{Knowledge Sheaves: \\ A Sheaf-Theoretic Framework for Knowledge Graph Embedding}

\aistatsauthor{ Thomas Gebhart \And Jakob Hansen \And  Paul Schrater }

\aistatsaddress{ University of Minnesota\\ Department of Computer Science \And  BlueLightAI, Inc. \And University of Minnesota\\ Department of Computer Science} ]

\begin{abstract}
  Knowledge graph embedding involves learning representations of entities---the vertices of the graph---and relations---the edges of the graph---such that the resulting representations encode the known factual information represented by the knowledge graph and can be used in the inference of new relations. We show that knowledge graph embedding is naturally expressed in the topological and categorical language of \textit{cellular sheaves}: a knowledge graph embedding can be described as an approximate global section of an appropriate \textit{knowledge sheaf} over the graph, with consistency constraints induced by the knowledge graph's schema. 
  This approach provides a generalized framework for reasoning about knowledge graph embedding models and allows for the expression of a wide range of prior constraints on embeddings.
  Further, the resulting embeddings can be easily adapted for reasoning over composite relations without special training.
  We implement these ideas to highlight the benefits of the extensions inspired by this new perspective.
\end{abstract}

\section{INTRODUCTION}


Knowledge graphs are structured knowledge bases which encode information about entities and their relationships. Each graph is a set of triplets---two entities and a relation between them---that represent facts about a domain. Modern knowledge graphs may contain millions of entities and billions of relational facts. As a result, efficiently representing knowledge graphs in a way amenable to large-scale inferential analysis is of great practical importance in areas such as biomedicine~\citep{santos2022knowledge, Hoyt2022Bioregistry}, content recommendation~\citep{sun2019research, guo2020survey}, and scientific discovery~\citep{sinha2015overview, auer2018towards}.

Knowledge graph embedding (KGE) has emerged as an important approach to encoding this type of data. In this approach, a representation of each entity and relation is learned, enabling tasks like knowledge graph completion, multi-hop reasoning, and other forms of inference \citep{chen2020review, ji2020survey}. 
A wide variety of knowledge graph embedding approaches exist \citep{bordes2013translating, trouillon2017complex, nickel2016holographic, ji2016knowledge, wang2014knowledge, lin2017learning, socher2013reasoning, jenatton2012latent, bordes2011learning, zhang2019quaternion}, each with its own motivation and theoretical framework.

Various taxonomies have been proposed to classify these approaches \citep{ji2020survey}, but they remain far from exhausting the space of possibilities. Describing and enforcing priors about the entities and relations by embedding in more complex spaces has attracted recent interest \citep{zhang2019quaternion,sun2019rotate,ebisu2017toruse}, but a general framework for these types of constraints is still lacking. We express the knowledge graph embedding task in a formal framework that illuminates the structural similarity across a variety of prior methods while also inspiring new possibilities for this family of embedding techniques.

Our approach stems from an algebro-topological perspective, using  cellular sheaves~\citep{curry_sheaves_2014} and their Laplacians~\citep{hansen_toward_2019} to construct a framework for knowledge graph embedding. Many of the goals of KGE---local agreement across relations, global consistency, typed representations, and multi-hop reasoning---can be readily formalized and interpreted in this sheaf-theoretic language.
A number of well-known KGE models may be subsumed within this framework, providing a uniform language for reasoning about the regularization and extension of existing models. This perspective also provides new benefits: the freedom to embed entities in spaces of varying dimension according to type semantics, control over symmetry or antisymmetry of relations, and a natural method (\emph{harmonic extension}) for reasoning over multi-hop queries. We implement these ideas and validate their performance on a range of complex query structures within two benchmark datasets, showcasing a theoretically-sound method for adapting knowledge graph embeddings trained on simple knowledge graph completion to more complex queries. 

\section{KNOWLEDGE GRAPHS}
A knowledge graph is often vaguely defined as a set of entities $\mathcal{E}$ together with a set of relations $\mathcal{R}$ between these entities. To facilitate the connection with cellular sheaves, we provide a more formal definition.
\begin{definition}\label{def:knowledge_database_schema}
Let $\mathcal{S}$ be a set of entity types, and $\mathcal{R}$ a set of relations. Suppose that each relation $r \in \mathcal{R}$ may hold between an entity of type $\mathfrak{h}(r) \in \mathcal{S}$ and an entity of type $\mathfrak{t}(r) \in \mathcal{S}$. The tuple $\mathcal{Q} = (\mathcal{S},\mathcal{R},\mathfrak{h},\mathfrak{t})$ is a \emph{knowledge database schema}. 
\end{definition}
Recall that a directed multigraph consists of a set $V$ of vertices and a set $E$ of edges, with two functions $h, t: E \to V$, where an edge $e \in E$ is thought of as going from $h(e)$ to $t(e)$. Note that under this definition a knowledge database schema is simply a directed multigraph, where each entity type is a vertex and each relation is an edge.

A knowledge graph then instantiates a schema in the form of a set of factual triplets which respect the typing from $\mathcal{Q}$:
\begin{definition}\label{def:knowledge_database}
Given a knowledge database schema $\mathcal{Q} = (\mathcal{S},\mathcal{R},\mathfrak{h},\mathfrak{t})$, a set $\mathcal{E}$ of entities, and a labeling function $s: \mathcal{E} \rightarrow \mathcal{S}$ which gives the type of each entity, a \emph{knowledge graph $G$ instantiating $\mathcal{Q}$} is a directed graph with vertex set $\mathcal{E}$ and edges $\mathcal{T} \subseteq \mathcal{E} \times \mathcal{R} \times \mathcal{E}$ whose elements $(h,r,t) \in \mathcal{T}$ must satisfy the type consistency conditions $\mathfrak{h}(r) = s(h)$ and $\mathfrak{t}(r) = s(t)$. 
\end{definition}
The relationship between the knowledge graph $G$ and its schema $\mathcal{Q}$ is captured by a \textit{graph morphism}.

\begin{definition}\label{def:graph_morphism}
Let $G$ and $H$ be directed multigraphs. A \emph{graph morphism}\footnote{It should be noted that there are other, slightly different definitions of this term; we have chosen this one because it appropriately models the structure of knowledge graphs.} $k: G \to H$ consists of a pair of functions $k_v: V(G) \to V(H)$ and $k_e: E(G) \to E(H)$ such that, for every edge $e$ of $G$, $h(k_e(e)) = k_v(h(e))$ and $t(k_e(e)) = k_v(t(e))$. That is, $k$ maps nodes to nodes and edges to edges in a way that respects the incidence relation.
\end{definition}
The type consistency condition on $G$ is precisely the requirement that the obvious map $G \to \mathcal{Q}$ be a graph morphism. For clarity, a simple example of a knowledge graph with schema is sketched in the appendix.

The schema $\mathcal{Q}$ and graph morphism $k$ are often given externally and can be quite simplistic, so it is common to refer to $G$ itself as the knowledge graph without reference to the other type-enforcing structures. 
Indeed, many popular benchmark datasets for knowledge graph embedding assume only one entity type (one vertex in $\mathcal{Q}$) with a collection of relations mapping this type to itself. This typing scheme is often chosen for convenience and may deserve further consideration if one wishes to embed hierarchical or other type-specific information within knowledge graph embeddings~\citep{jain2021do}. 

\subsection{Knowledge Graph Embedding}

The problem of knowledge graph embedding is, broadly speaking, that of finding representations $\vec{x}_h$ for each entity $h$ and representations $\mat{R}_r$ for each relation type $r$ such that the truth value of the tuple $(h, r, t)$ 
may be recovered from $(\vec{x}_h, \mat{R}_r, \vec{x}_t)$. Note that the entities being embedded are the vertices of $G$, while the relations being embedded are the edges of $\mathcal{Q}$; the edges whose existence we want to be able to infer are the edges of $G$. A common inferential goal for a knowledge graph embedding is to predict the truth value of new triples from their learned representations alone. The predicted truth value of such a relation is typically determined from the output of some scoring function which takes a vectorized triplet to a real number representing the model's degree of confidence in its truth.

The classic Structured Embedding model \citep{bordes2011learning} represents each entity as a vector $\vec{x}_e \in \mathbb{R}^d$ and each relation $r$ as a pair of $d \times d$ matrices $(\mat{R}_{rh}, \mat{R}_{rt})$,
and uses the scoring function $f^{SE}(h,r,t) = \|\mat{R}_{rh}\vec{x}_h - \mat{R}_{rt}\vec{x}_t\|^2$. That is, a relation embedding provides a pair of linear transformations applied to the head and tail entity embeddings before comparing them. This model is a motivating example for many other embedding methods, and is a good starting point for the generalization to sheaf embedding models.

\section{CELLULAR SHEAVES}

Abstractly, a sheaf is a mathematical object which tracks the assignment of data to open sets of a topological space. While sheaf theory has existed for nearly a century, only in the past decade has a discretized and computationally tractable theory of \textit{cellular sheaves} received significant attention and development~\citep{curry_sheaves_2014}. While the theory behind these structures can become quite intricate, all the machinery we will need can be explained using basic graph theory and linear algebra. Our introduction here will be specifically adapted to the motivating example of knowledge graphs.

\begin{definition}\label{def:cellular_sheaf}
A \emph{cellular sheaf} $\Fc$ on a directed graph $G = (V,E)$ consists of the following data:
\begin{itemize}
    \item a vector space $\Fc(v)$ for each vertex $v \in V$ of $G$
    \item a vector space $\Fc(e)$ for each edge $e \in E$ of $G$,  
    \item a linear transformation $\mat{\Fc}_{v \face_h e} : \Fc(v) \to \Fc(e)$ for each pair $(v,e)$ with $h(e) = v$, and
    \item a linear transformation $\mat{\Fc}_{v \face_t e}: \Fc(v) \to \Fc(e)$ for each pair $(v, e)$ with $t(e) = v$.
\end{itemize}
\end{definition}
The sheaf structure over a graph associates a space of data, called a \emph{stalk}, to each node and edge. We refer to the linear maps between stalks as \emph{restriction maps}.
For most pairs $(v,e)$, at most one of the restriction maps $\mat{\Fc}_{v \face_h e}$ and $\mat{\Fc}_{v \face_t e}$ can exist, in which case we simplify the notation to $\mat{\Fc}_{v \face e}$. The only exception is when $e$ is a self-loop on the vertex $v$, when both maps exist and may be distinct.

If $G$ is a knowledge graph, we can think of each vertex stalk $\Fc(v)$ as a vector space from which we can choose a representation of the entity $v$.  The restriction maps $\Fc_{v \face e}$ are used to compare entity embeddings with respect to an edge representing a particular relation, and the edge stalks $\Fc(e)$ are the spaces in which this comparison happens.

Another way to say this is that the restriction maps of $\Fc$ encode consistency constraints for entity embeddings. 
For an edge $e$ between vertices $u$ and $v$, we say that a choice of two embeddings $\vec{x}_v \in \Fc(v)$, $\vec{x}_u \in \Fc(u)$ is \emph{consistent over $e$} if $\mat{\Fc}_{v \face e} \vec{x}_v = \mat{\Fc}_{u \face e} \vec{x}_u$. 
In Structured Embedding, all stalks are the same $\mathbb{R}^d$, and the relation embedding matrices become the restriction maps. The score for a relation is zero if the entity embeddings are consistent over that edge.

The space of all possible entity embeddings (i.e., a choice of an embedding vector for each entity) is known as the space of $0$-cochains of $\Fc$, and is denoted $C^0(G;\Fc)$. Each edge of $G$ imposes a constraint on $C^0(G;\Fc)$ by requiring the data over its two incident vertices to be consistent over the edge. The subspace of $H^0(G;\Fc) \subseteq C^0(G;\Fc)$ consisting of cochains that satisfy all these local constraints is called the space of \emph{global sections} of $\Fc$.

Similarly, the space of all choices of one vector in each edge stalk $\Fc(e)$ is the space of $1$-cochains $C^1(G;\Fc)$. 
The space of global sections $H^0(G;\Fc)$ is the nullspace of a linear transformation $\mat{\delta}: C^0(G;\Fc) \to C^1(G;\Fc)$ whose output on an edge $e: u \to v$ is given by the formula
\[(\mat{\delta} \vec{x})_e = \mat{\Fc}_{v \face e} \vec{x}_v - \mat{\Fc}_{u \face e} \vec{x}_u\]
where the edge $e$ is directed $u \to v$. Therefore, if $\mat{\delta}\vec{x} = 0$, then $\vec{\Fc}_{v \face e} \vec{x}_v = \vec{\Fc}_{u \face e} \vec{x}_u$ for every edge $e = u \sim v$. From the coboundary operator we may construct the \emph{sheaf Laplacian} $\mat{L}_\Fc = \mat{\delta}^T\mat{\delta}$ \citep{hansen_toward_2019}.

The sheaf Laplacian provides a continuous measure of consistency for potential entity embeddings. As the elements of $H^0(G;\Fc)$ represent globally consistent choices of data on vertices, we can use the sheaf Laplacian to quantify how close any data assignment in $C^0(G;\Fc)$ is to consistency: 
\begin{equation}\label{eq:quadratic_form}
    \vec{x}^T \mat{L}_\Fc \vec{x} = \sum\limits_{e = u \sim v \in E} \|\mat{\Fc}_{u \face e}\vec{x}_u - \mat{\Fc}_{v \face e} \vec{x}_v \|^2.
\end{equation}
The closer $\vec{x}^T\mat{L}_\Fc \vec{x}$ is to zero, the closer the section $\vec{x}$ is to consistency, and when $\vec{x}^T\mat{L}_\Fc \vec{x} = 0$, $\vec{x}$ is fully consistent.

Note that each individual term in this sum is of the same form as the Structured Embedding scoring function.
As we will observe, a large family of knowledge graph embedding methods implicitly target similar notions of global consistency in the optimization of entity and relation embeddings derived from knowledge graph triplets. 

To fully exploit the flexibility of this sheaf-theoretic perspective, we introduce one more tool which allows us to transfer a sheaf from one graph to another. This operation will depend on a graph morphism as described in Definition~\ref{def:graph_morphism}. 
\begin{definition}\label{def:pullback}
Given multigraphs $G$ and $H$, a graph morphism $k: G \rightarrow H$ sending nodes to nodes and edges to edges, and $\Fc$ a sheaf on $H$, the \emph{pullback sheaf} $k^*\Fc$ is a sheaf on $G$ with stalks $k^*\Fc(\sigma) = \Fc(k(\sigma))$ and restriction maps $\mat{k^*\Fc}_{v \face_\bullet e} = \mat{\Fc}_{k(v) \face_\bullet k(e)}$. 
\end{definition}
This operation replicates the local structure of $\Fc$ on all the parts of $G$ that map to the same part of $H$. Note that, for instance, this allows us to restrict a sheaf on $G$ to any subgraph by using the pullback over the inclusion morphism.
Cochains of $\Fc$ may also be pulled back to cochains of $k^*\Fc$ by a similar process. If $\vec{x} \in C^i(H;\Fc^H)$, we define $k^*\vec{x} \in C^i(G;\Fc)$ by $(k^*\vec{x})_\sigma = \vec{x}_{k(\sigma)}$. 
It is straightforward to show that if $\vec{x} \in H^0(H;\Fc)$, then $k^*\vec{x} \in H^0(G;k^*\Fc)$; that is, global sections of the initial sheaf automatically become sections of its pullback.

\section{KNOWLEDGE SHEAVES AND EMBEDDINGS}\label{sec:sheaf_learning}
We are now ready to define sheaf embeddings of knowledge graphs, using the concepts introduced above. This approach separates relation embeddings from entity embeddings: relation embeddings are sheaves on the schema graph $\mathcal{Q}$, while entity embeddings are $0$-cochains of the pullback of this sheaf to the knowledge graph $G$. More formally:
\begin{definition}\label{def:knowledge_sheaf}
Given a knowledge database schema $\mathcal Q = (\mathcal{S},\mathcal{R},\mathfrak{h},\mathfrak{t})$, a \emph{knowledge sheaf} $\Fc$ modeled on $\mathcal Q$ corresponds to a choice of vertex stalk spaces $\Fc(s)$ for each entity type $s \in \mathcal{S}$, edge stalk spaces $\Fc(r)$ for each relation type $r \in \mathcal{R}$, and linear maps $\mat{\Fc}_{h \face_h r}: \Fc(\mathfrak{h}(r)) \to \Fc(r)$ and $\mat{\Fc}_{t \face_t r}: \Fc(\mathfrak{t}(r)) \to \Fc(r)$ for each $r \in \mathcal{R}$.
\end{definition} That is, a knowledge sheaf is simply a cellular sheaf on the directed multigraph $\mathcal{Q}$. To make the space of knowledge sheaves on $\mathcal{Q}$ into a vector space, we assign a dimension $d_s$ for the stalk over each vertex (entity type) $s$ and a dimension $d_r$ for the stalk over each edge (relation type) $r$. This gives a generalized version of Structured Embedding: a relation $r$ that may hold between entities of type $s, t$ is represented by the two restriction maps $\mat{\Fc_{s \face r}}$ and $\mat{\Fc_{t \face r}}$, which are matrices of shapes $d_r \times d_s$ and $d_r \times d_t$.

To produce the space of entity embeddings for a knowledge graph corresponding to schema $\mathcal{Q}$, we use a pullback of a knowledge sheaf.
\begin{definition}\label{def:sheaf_embedding}
Given a graph morphism $k: G \to \mathcal{Q}$ instantiating a knowledge graph $G$ from a schema $\mathcal{Q}$, a \emph{sheaf embedding} of $G$ is a knowledge sheaf $\Fc$ on $\mathcal{Q}$ together with a $0$-cochain $\vec{x} \in C^0(G;k^*\Fc)$.
\end{definition} While this definition depends on the sheaf $\Fc$, the resulting space of possible entity embeddings depends only on the choice of dimension for each stalk of $\Fc$. The embedding of an entity of type $s$ is simply a $d_s$-dimensional vector; if there are $N$ entities, their embeddings combine by concatenation into an $Nd_s$-dimensional vector, which is an element of $C^0(G;k^*\Fc)$.


For convenience and clarity, we will also denote the sheaf $k^*\Fc$ by $\Fc^G$. Note that if $H \subseteq G$ is a subgraph, we can restrict the morphism $k$ to the vertices and edges in $H$ and obtain a morphism $k_H: H \to \mathcal{Q}$. We will denote the pullback $k_H^* \Fc$ by $\Fc^H$. The restriction of a 0-cochain $\vec{x}$ of $k^* \Fc$ to the nodes in $H$ will be denoted $\vec{x}^H$. 

A concrete description of $\Fc^G$ is as follows:
For an entity $v$, $\Fc^G(v) = \Fc(s(v))$, and for an edge $e = (h,r,t)$, $\Fc^G(e) = \Fc(r)$. The restriction maps for $e$ are given by $\mat{\Fc}^G_{h \face e} = \mat{\Fc}_{h \face r_e}$ and $\mat{\Fc}^G_{t \face e} = \mat{\Fc}_{t \face r_e}$. 

It is important to observe that not every sheaf on $G$ arises as a knowledge sheaf. The vertex stalk spaces and edge stalk spaces are determined by the types of entities and relations, respectively, and the restriction maps for an edge are determined entirely by the relation described by that edge. Since many edges in $G$ correspond to the same relation type, this is a form of parameter sharing and greatly reduces the complexity of the knowledge graph embedding.

\subsection{Consistent and Contrastive Embeddings}
Definition~\ref{def:sheaf_embedding} specifies the spaces of entity and relation embeddings, but does not capture the desired representational power of the embedding. In most knowledge graph representation literature, this is done by specifying a loss function; we prefer to specify the desired outcome and then construct a corresponding loss function. 

\begin{definition}\label{def:consistent_sheaf_embedding}
Let $k: G \to \mathcal{Q}$ be a knowledge graph with schema $\mathcal{Q}$. A \emph{consistent sheaf embedding} of $G$ is a knowledge sheaf $\Fc$ on $\mathcal{Q}$ together with a section $\vec{x} \in H^0(G;k^*\Fc)$.
\end{definition}
That is, a consistent sheaf embedding is one where 
embeddings of true relational facts are consistent as measured by the sheaf restriction maps; if $(h,r,t)$ is a true relation, then $\mat{\Fc^G_{h \face r}} \vec{x}_h = \mat{\Fc^G_{t \face r}} \vec{x}_t$. However, this definition does not ensure we can distinguish true from false triples. A trivial entity embedding with $\vec{x}_{e} = 0$ for every entity $e$ would be consistent, but useless. To distinguish true from false relations, we need negative examples, which can be seen as forming a new knowledge graph with the same schema. To capture this requirement, we make a new definition:


\begin{definition}\label{def:contrastive_sheaf_embedding}
Let $k: G \to \mathcal{Q}$ and $\tilde{k}: \tilde{G} \to \mathcal{Q}$ be knowledge graphs with the same schema and vertex sets. Call $G$ the \emph{positive} knowledge graph, containing relations known to be true, and $\tilde{G}$ the \emph{negative} knowledge graph of  triples assumed to be false. Let $\tilde{\mathcal A}$ be a collection of subgraphs of $\tilde{G}$.
A \emph{contrastive sheaf embedding} of $G$ with respect to $\tilde{G}$ and $\tilde{\mathcal{A}}$ consists of a consistent sheaf embedding of $G$ such that
for every $\tilde{H} \in \tilde{\mathcal{A}}$, $\vec{x}^{\tilde{H}}$ is not a section of $\Fc^{\tilde H}$.
\end{definition}

In other words, the entity and relation embeddings are consistent for every relation in $G$, and inconsistent for every selected subgraph of $\tilde{G}$. 
Thus, if $\tilde{\mathcal{A}}$ is the set of all subgraphs of $\tilde{G}$ with two vertices and one edge, a contrastive sheaf embedding will be able to distinguish perfectly between relations in $G$ and relations in $\tilde{G}$ by checking if $\vec{x}$ is consistent over the relevant edge or not. 

The use of negative examples to constrain knowledge graph embeddings is nearly universal in the literature, and there are various standard choices for constructing $\tilde{G}$ \citep{ali2020benchmarking}. Taking $\tilde{G}$ to be the complement of $G$ relative to $\mathcal{Q}$ corresponds with the \emph{closed world assumption}: all relations not known to be true must be false. By contrast, the \emph{open world assumption} corresponds to a $\tilde{G}$ with no edges at all, returning us to Definition~\ref{def:consistent_sheaf_embedding}. These extremes are mostly unsuitable for learning embeddings, and intermediate assumptions like the \emph{local closed world assumption} are typically used. This corresponds to constructing $\tilde{G} \to \mathcal{Q}$ by taking the node set of $G$, and creating an edge $\tilde{e} = u \to v'$ with $\tilde{k}(\tilde{e}) = r$ if there exists a $v \neq v'$ and an edge $e = u \to v$ in $G$ with $k(e) = r$. 

In practice, it is often difficult to find an exact section of a sheaf, if one even exists, so we need versions of these definitions that replace the strict equality constraints for consistency with inequalities. This is where scoring functions come into play.

\begin{definition}\label{def:sheaf_scoring}
A \emph{sheaf scoring function} is a function $V$ taking a graph $G$, a sheaf $\Fc$ on $G$, and a 0-cochain $\vec{x}$ of $\Fc$, such that $V_{G,\Fc}(\vec{x}) \geq 0$, and $V_{G,\Fc}(\vec{x}) = 0$ exactly when $\vec{x}$ is a section of $\Fc$.
\end{definition}
The canonical example is the Laplacian quadratic form $V_{G,\Fc}(\vec{x}) = \vec{x}^T \mat{L}_\Fc \vec{x}$, which we will use almost exclusively, but there are many other options. For instance, any norm induces a scoring function $V_{G,\Fc}(\vec{x}) = \|\mat{\delta} \vec{x}\|$. Note that many sheaf scoring functions (including the Laplacian quadratic form) can be decomposed into a sum with one term for each edge in $G$, but this is not required to be the case. 

\begin{definition}\label{def:gapped_sheaf_embedding}
Given a sheaf scoring function $V$, a margin $\gamma \geq 0$, positive and negative knowledge graphs $k, \tilde{k}: G, \tilde{G} \to \mathcal{Q}$, and a set $\mathcal{A}$ of pairs $(H, \tilde{H})$ of subgraphs of $G, \tilde{G}$, a \emph{$\gamma$-gapped contrastive sheaf embedding} is a sheaf embedding of $G$ such that 
     for every pair of subgraphs $(H, \tilde{H}) \in \mathcal{A}$, $V_{\tilde{H},\Fc^{\tilde{H}}}(\vec{x}^{\tilde{H}}) - V_{H, \Fc^{H}}(\vec{x}^{H}) > \gamma$.
\end{definition}

A common choice for the set of contrastive pairs $\mathcal{A}$ is to choose all pairs $(H, \tilde{H})$, where both graphs consist of a single edge with the same pair of incident vertices. 
Note that for any $\gamma$, a gapped contrastive sheaf embedding with properly chosen pairs of contrastive subgraphs still enables us to perfectly distinguish between relations in $G$ and relations in $\tilde{G}$. However, this relaxation makes it easier to quantify an embedding's degree of consistency, and to produce tractable objective functions for training.
Indeed, Definition~\ref{def:gapped_sheaf_embedding} leads directly to the \emph{margin ranking loss} for learning knowledge graph embeddings. The decomposition of the criterion over the pairs of subgraphs in $\mathcal{A}$ also provides a natural way to construct training batches~\citep{schlichtkrull2018modeling}.

It should be noted that an embedding satisfying Definition~\ref{def:gapped_sheaf_embedding} only guarantees the ability to distinguish between relations in $G$ and relations in $\tilde{G}$ by comparison with each other. That is, given two relations, with a guarantee that one is in $G$ and the other is in $\tilde{G}$, we can determine which is which by comparing the corresponding scores. However, given a single relation, there is no way to identify whether it lies in $G$ or $\tilde{G}$. Further criteria could be added
to address this point, but we will focus on the purely contrastive case, as it is used in most popular knowledge graph embedding approaches.

\subsubsection{Translational Embeddings}
One perspective on Definition~\ref{def:consistent_sheaf_embedding} is that it asks for a vector $\vec{x}$ and a linear map $\mat{\delta}$ with particular structure such that $\mat{\delta} \vec{x} = 0$. We may ask whether the zero vector should be special---why not also learn a vector $\vec{y}$ such that $\mat{\delta} \vec{x} = \vec{y}$? This turns out to be a bit too much freedom, since for any $\mat{\delta}$ and $\vec{x}$ we could just choose $\vec{y} = \mat{\delta} \vec{x}$, but this restriction makes sense if we require $\vec{y}$ to be the pullback $k^* \vec{z}$ of a 1-cochain of $\Fc$ on $\mathcal{Q}$. This amounts to requiring, for every edge $e: u \to v$ in $G$ over a relation $r: h \to t$ in $\mathcal{Q}$, that $\mat{\Fc}_{h \face r} \vec{x}_u - \mat{\Fc}_{t \face r} \vec{x}_v = \vec{z}_r$. We call this a \emph{translational sheaf embedding}, as it requires the embeddings of entities to agree with each other after a translation in the edge stalk.

A consistent translational sheaf embedding exists precisely when a standard sheaf embedding exists. This can be seen by noting that if $k^*\vec{z}$ is in the image of $\mat{\delta}_{k^*\Fc}$, then we can subtract any preimage from $\vec{x}$ to get an entity embedding for which $\mat{\delta}_{\Fc} \vec{x} = 0$. However, once we add negative constraints on the embedding, the picture is not so simple, and a nonzero target 1-cochain may be useful. 
There are natural generalizations of the previous definitions to include a translational component, and some extensions are described in the appendix.

\subsection{Loss Functions}
We have now specified embedding spaces for entities and relations, as well as consistency conditions for embeddings. To learn an embedding, we construct a loss function defined on these embedding spaces whose minima correspond to embeddings satisfying the conditions in one of the definitions~\ref{def:sheaf_embedding}--\ref{def:gapped_sheaf_embedding}. 

For instance, the Laplacian quadratic form $\vec{x}^T \mat{L}_{\Fc^G}\vec{x}$ attains its minimum value precisely when $\vec{x}$ and $\Fc$ form an embedding according to definition~\ref{def:consistent_sheaf_embedding}. Note that since both $\mat{L}_{\Fc^G}$ and $\vec{x}$ depend on the embedding parameters, this loss function is not simply a convex quadratic function. 

Knowledge graph embedding objective functions are typically thought of as being constructed from the scoring function that evaluates the quality of the embedding for a single edge of $G$. This is a special case of the scoring functions of Definition~\ref{def:sheaf_scoring}, which evaluates an embedding on an arbitrary subgraph of $G$. As noted above, the Laplacian quadratic form $\vec{x}^T \mat{L}_{\Fc^G}\vec{x}$ is the aggregate sum over all edges of the scoring function used in the Structured Embedding model; in sheaf-theoretic notation this is
\begin{equation}\label{eq:scoring_function}
f^{\text{SE}}(h,r,t) = \|\mat{\Fc}_{h \face r}\vec{x}_h - \mat{\Fc}_{t \face r} \vec{x}_t\|^2.
\end{equation}

Following our discussion of translational embeddings in the previous section, we may define a translational scoring function similarly:
\begin{equation}\label{eq:translational_scoring_function}
    f^{\text{TransX}}(h, r, t) = \|\mat{\Fc}_{h \face r}\vec{x}_h + \vec{r}_r - \mat{\Fc}_{t \face r} \vec{x}_t\|^2.
\end{equation} The scoring function $f^{\text{TransX}}$ is equivalent to the TransR \citep{lin2015learning} scoring function, and when $\mat{\Fc}_{h \face r} = \mat{\Fc}_{t \face r_r} = \mat{I}$, this scoring function is equivalent to that of TransE \citep{bordes2013translating}.

The Laplacian quadratic form does not incorporate any negative information about our knowledge graph. Knowledge graph embedding techniques typically construct an objective by evaluating the chosen scoring function on contrastive pairs in such a way as to encourage the score of the true relation to be smaller, implying better consistency. The contrastive embedding definitions given above are meant to capture this notion. In particular, Definition~\ref{def:gapped_sheaf_embedding} lends itself nicely to an objective function.
To learn a $\gamma$-gapped sheaf embedding of our knowledge graph, we use the \emph{margin ranking loss}:
\begin{equation}\label{eq:loss_marginranking}
    \mathcal{L}_{m} = \sum\limits_{(H,\tilde{H}) \in \mathcal{A}} \max(0, V_{H, \Fc^H}(\vec{x}^H) + \gamma - V_{\tilde{H}, \Fc^{\tilde{H}}}(\vec{x}^{\tilde{H}}))
\end{equation} which is equal to zero if and only if $\Fc$ and $\vec{x}$ form a $\gamma$-gapped contrastive sheaf embedding with respect to the contrastive graph pairs given in $\mathcal{A}$.

\subsection{Learning Multiple Sections}\label{sec:learning_multiple_sections}

It is desirable to produce knowledge graph embeddings which encode knowledge in a robust and generalized manner so that these embeddings may be applicable to downstream knowledge tasks involving unseen data. From the purview of sheaf embedding, one way to coerce these knowledge graph representations to be more general is to force the space of approximate sections of the learned knowledge sheaves to be large. In other words, we would like $\vec{x}^T\mat{L}_{\Fc^G}\vec{x}$ to be small for as many choices of $0$-cochain $\vec{x}$ as possible. Up to this point, we have defined a sheaf embedding as consisting of a single $0$-cochain $\vec{x} \in C^0(G;\Fc^G)$ and, in the translational case, $1$-cochain $\vec{r} \in C^1(G;\Fc^G)$ that represent the entity and relation embeddings, respectively learned from the training data. One way to improve the robustness of our sheaf embedding is to learn multiple $0$-cochains simultaneously, which can be thought of as an ensemble learning approach that may help mitigate errors due to initialization, sampling, and labeling~\citep{adlam2020understanding}. Learning a set of $k$ independent cochains is simple to implement: instead of learning a single $d_v$-dimensional vector $\vec{x}_v$ for each entity, we learn a $d_v \times k$ matrix $\mat{X}_v$; the loss function is updated accordingly by using the Frobenius norm on matrices. It is important to note that the relation embeddings do not increase in size, which may help avoid some types of overfitting.



\subsection{Modeling Knowledge Priors}
The choice of representational prior has significant ramifications for the performance of knowledge graph embedding models~\citep{sun2019rotate,zhang2019quaternion,cai2019group,patel2021modeling}.
In addition to generalizing a number of popular knowledge graph embedding approaches, this knowledge sheaf framework helps clarify the options for constraining knowledge graph embeddings to better capture the semantics of the underlying knowledge domain. 
The structure of the restriction maps $\mat{\Fc}_{\bullet \face r}$ for each relation $r$ provides control for modeling symmetric, asymmetric, one-to-many, many-to-one, or one-to-one relations by choosing the proper structure for the restriction maps across each edge type in $\mathcal{R}$. For example, a symmetric relationship may be enforced by requiring that $\mat{\Fc}_{h \face r} = \mat{\Fc}_{t \face r}$. The choice of edge stalk space $\Fc(r)$ for each relation type $r$ provides flexibility for determining the space within which entity embeddings are compared across incident edges. For example, setting $\dim \Fc(r) < \dim \Fc(h)$ means that an entity embedding $\vec{x}_h$ can be consistently extended across $r$ to many different embeddings $\vec{x}_t$. 

The linear transformations represented in the restriction maps can also be constrained to regularize the learned embeddings: forcing $\mat{\Fc}_{\bullet \face r}$ to be orthogonal requires entity embeddings to be comparable as a rotation across $r$. We experiment with these latter two parametrizations in Section~\ref{sec:experiments}. Finally, when the schema $\mathcal{Q}$ has multiple entity types, the embedding stalk space can vary across these types, decreasing parameterization for types which can be modeled using few dimensions.

\subsection{Inference with Sheaf Embeddings}

The standard knowledge graph completion task involves finding pairs of entities linked by a given relationship which is not already encoded in the knowledge graph. The standard approach to this task involves ranking potential relations using the scoring function defined by the embedding model, and this approach applies equally well to sheaf embeddings using a sheaf scoring function. The advantage of reframing knowledge graph embedding as learning a knowledge sheaf is that one can exploit the spectral properties of cellular sheaves to naturally extend these embeddings to answer more complex queries.

\subsubsection{Multi-Hop Reasoning}
Complex relationships may be constructed by composing the basic relations of a knowledge graph. 
Often the resulting relationship is one of the basic relation types represented in the knowledge graph, but it is also possible to construct more complex relations by composition. 
For instance, the relations ``$x$ is a child of $y$'' and ``$y$ is a child of $z$'' compose to ``$x$ is a grandchild of $z$.'' The term ``multi-hop reasoning'' is often used for deductions of this sort~\citep{guu2015traversing, gardner2014incorporating, toutanova2016compositional}. 
The sheaf Laplacian provides a natural tool for addressing these composite queries.
We describe here the construction for non-translational sheaf embeddings; the extension to translational embeddings is detailed in the appendix.

If we wish to infer the possible endpoint of a sequence of relations $r_1; r_2;\cdots;r_k$, beginning at known entity $u_0$ and ending at some to-be-determined entity $u_k$, we can construct a chain of edges with these relations, and optimize for their combined discrepancy. That is, we consider a new knowledge graph $H$ modeled on $\mathcal{Q}$ with vertex set $v_0,\ldots,v_k$. The knowledge sheaf $\Fc$ also induces a sheaf $\Fc^H$ on $H$ as before. 
If we match each vertex of $H$ with a vertex of $G$, 
(in a way consistent with the schema), 
the entity embeddings for $G$ give us a $0$-cochain of $\Fc^H$. We want to find the matching that makes this $0$-cochain as close to a section as possible. 
The corresponding optimization problem can be expressed as 
\begin{equation}\label{eq:entitychaining}
    \argmin_{u_1,\ldots,u_k \in \mathcal{E}} \sum_{i=1}^k \|\mat{\Fc}^H_{v_{i-1} \face e_i} \vec{x}_{u_{i-1}} - \mat{\Fc}^H_{v_i \face e_i}  \vec{x}_{u_{i}}\|^2.
\end{equation} 
Naively, finding the best fit for a chain of length $k$ requires evaluating the objective function at $\vert\mathcal{E}\vert^k$ tuples of entities. 
Other approaches to this problem try to find approximate solutions, e.g. by simply greedily extending to the best entity at each step or ignoring the interior nodes altogether and constructing some joint composite relation, thus simplifying to single-hop graph completion~\citep{lin2018multi,guu2015traversing}. 

We offer a new approach based on a related optimization problem. Rather than choosing one entity from the knowledge database for each intervening node $u_1,\ldots,u_{k-1}$, we optimize directly over the entity embedding space, with the intermediate cost function
\begin{equation}\label{eq:sheafcost}
    V(\vec{y}) = \sum_{i=1}^k \|\mat{\Fc}^H_{v_{i-1} \face e_i} \vec{y}_{i-1} - \mat{\Fc}^H_{v_i \face e_i}  \vec{y}_{i}\|^2 = \vec{y}^T \mat{L}_{\Fc^H} \vec{y}.
\end{equation}
This is a relaxation of~\eqref{eq:entitychaining}, as $\vec{y}_i$ need not be the embedding of a known entity. The relaxed problem of finding the best-fitting tail entity $u_k$ for the composite relation is then
\begin{equation}\label{eq:harmonicextension}
    \argmin_{u_k\in \mathcal{E}}\; \left(\min_{\vec{y} \in C^0(H;\Fc^H)} V(\vec{y})\;\text{s.t. } \vec{y}_0 = \vec{x}_{u_0}, \; \vec{y}_{k} = \vec{x}_{u_k}\right).
\end{equation}


The inner optimization problem, depending on $\vec{x}_{u_0}$ and $\vec{x}_{u_k}$, is the problem of \emph{harmonic extension} of a 0-cochain defined on a boundary subset of vertices $B$, which here is $\{v_0,v_k\}$. 
This problem is convex and quadratic, so the optimal value is unique, but the optimizer may not be. 
A brief argument using Lagrange multipliers shows that an equivalent problem is to find a 0-cochain $\vec{y} \in C^0(H;\Fc^H)$ such that $\vec{y}_{0} = \vec{x}_{u_0}$, $\vec{y}_k = \vec{x}_{u_k}$, and $\mat{L}_{\Fc^H} \vec{y} = \vec{0}$ on nodes not in $B$. When there is a unique solution, its values on $U$, the complement of $B$, are given by the formula $\vec{y}_U = -\mat{L}[U,U]^{-1}\mat{L}[U,B]\vec{y}_B$, where $\vec{y}_B$ is determined by the embeddings $\vec{x}_{u_0}$ and $\vec{x}_{u_k}$, and we drop the subscript on $\mat{L}_{\Fc^H}$. Then the minimum value of the inner optimization problem in~\eqref{eq:harmonicextension} is
\[V(\vec{y}^*) = \vec{y}_B^T \left(\mat{L}[B,B] -\mat{L}[B,U]\mat{L}[U,U]^{-1} \mat{L}[U,B]\right)\vec{y}_B.\]


The matrix in this formula is the \emph{Schur complement} $\mat{L}/\mat{L}[U,U]$ of $\mat{L}[U,U]$ in $\mat{L}$.\footnote{When $\mat{L}[U,U]$ is not invertible, we may use its Moore--Penrose pseudoinverse $\mat{L}[U,U]^\dagger$.} We can think of it as defining a scoring function for the composite relation. In fact, by factorizing $\mat{L}/\mat{L}[U,U]$, it is possible to extract from this a pair of matrices that can be thought of as an induced embedding for the multi-hop relation.

\subsubsection{Complex Composite Relations}
\begin{figure}
    \centering
    \includegraphics[width=.9\columnwidth,keepaspectratio]{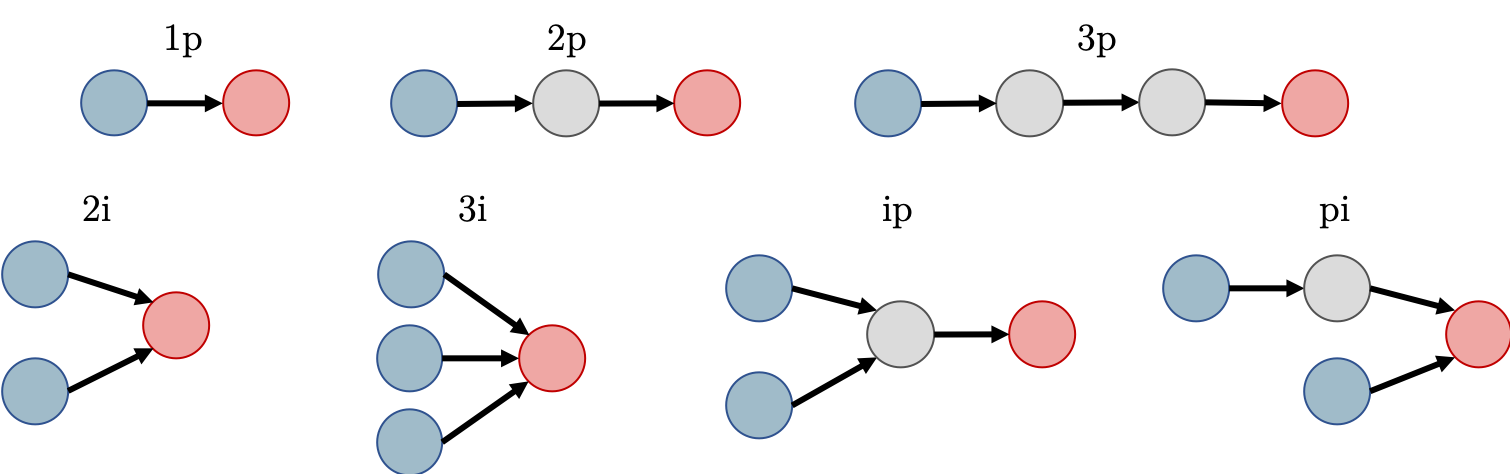}
    \caption{Illustration of complex query structures. Unknown entities are colored gray. Source entities are colored blue and target entities are colored red.}
    \label{fig:queries}
\end{figure}

We need not limit ourselves to composing relations in linear chains: harmonic extension adapts effortlessly to more complex networks of relations like those displayed in Figure~\ref{fig:queries}. Let $H$ be any knowledge graph with schema $\mathcal Q$. The learned knowledge sheaf extends to $H$ as before, and its sections over $H$ correspond to collections of entity embeddings jointly satisfying the relations. We construct a boundary set of vertices $B$ given by the entities of interest and denote its (possibly empty) complement by $U$, obtaining the Schur complement $\mat{L}_{\Fc^H}/\mat{L}_{\Fc^H}[U,U]$. The quadratic form $V(\vec{y}_B) = \vec{y}_B^T(\mat{L}_{\Fc^H}/\mat{L}_{\Fc^H}[U,U])\vec{y}_B$ finds the minimal value of a problem analogous to the inner problem in~\eqref{eq:harmonicextension}, constraining the values of $\vec{y}$ on $B$ to equal $\vec{y}_B$. We can then fix the values of $\vec{y}_B$ on some source subset of vertices $S$ to be equal to the embeddings of some given entities $\{u_s\}_{s\in S}$, and test the embeddings $\vec{x}_t$ for other entities $t$ to find the entities that minimize $V(\vec{y}_B)$ subject to $\vec{y}_S = \vec{x}_S$. 

For further insight regarding harmonic extension as a method for solving complex queries, it is helpful to note the relationship between the Schur complement and marginalization when entity embeddings are distributed as multivariate Gaussians~\citep{von2014mathematical}, as detailed in the appendix. Further, recall that sheaf embedding generalizes Structured Embedding, and with the addition of non-trivial $1$-cochains \eqref{eq:translational_scoring_function} represents a generalization of TransR. Harmonic extension provides a way to apply any of these models to multi-hop and complex composite queries in a theoretically justified manner which, to the authors' knowledge, is a first for models like Structured Embedding or TransR that are not purely translational or bilinear~\citep{guu2015traversing}. 



\section{EXPERIMENTS}\label{sec:experiments}

To validate the approach to answering complex queries detailed in the previous section and to highlight the flexibility of sheaf embedding, we compare the performance of a number of sheaf embedding models on two benchmark datasets: NELL-995~\citep{xiong2017deeppath} and FB15k-237~\citep{toutanova-etal-2015-representing}. These experiments are not intended to achieve state-of-the-art performance. Rather, our aim is to investigate the effect of model regularization choices detailed in Section~\ref{sec:sheaf_learning} and to showcase the extensibility of this framework to the evaluation of complex queries. We implement these models in the open-source knowledge graph embedding package Pykeen~\citep{ali2021pykeen}, allowing us to train and evaluate these models in a manner that is both reproducible and comparable to other embedding techniques. Models are trained and tested on an Nvidia GeForce GTX 1080 GPU with 8GB RAM. 

We train each model according to the traditional graph completion task, learning embeddings by minimizing the loss associated to scoring both true triplets and corrupted triplets (Equation~\ref{eq:loss_marginranking}). At test time, we evaluate each model on the dataset's held-out test set along with the 1p, 2p, 3p, 2i, 3i, ip, and pi complex query structures displayed in Figure~\ref{fig:queries} and originally detailed in~\citet{ren2020beta}. Performance is evaluated on both an ``easy''  and ``hard'' subset of complex query answers. The ``easy'' subset contains queries whose full structure is encountered within the training set whereas the ``hard'' subset requires at least one fact to be inferred that was not present in the training set. The query structures themselves correspond to logical multi-hop path query structures (*p), intersectional queries (*i) and a combination of the two (ip and pi). Entities are scored based on their solution to the associated optimization problem (\ref{eq:harmonicextension}). For each complex query type, model performance is evaluated according to two canonical measures: the mean reciprocal ranking (MRR) and the hits at 10 (H@10), both of which measure the relative ranking of the true solution to a query among the ordered score of all entities. Note that the models are only trained on the traditional triplet scoring task and not on the complex queries themselves. 

We train four types of models. The \verb|ExtensionSE| model implements scoring function \eqref{eq:scoring_function} which is a generalization of Structured Embedding~\citep{bordes2011learning}. To observe the effects of compressive comparison across relations, we vary the edge stalk space $\dim \Fc(r)$ and assume all edge types share this space for this model. We also train an \verb|ExtensionSE_orthogonal| model which applies an orthogonal parameterization to all restriction maps of the model. The \verb|ExtensionTranslational| model implements scoring function \eqref{eq:translational_scoring_function} while the \verb|ExtensionTransE| model implements the same scoring function but sets $\mat{\Fc}_{h \face r} = \mat{\Fc}_{t \face r} = \mat{I}$ which is equivalent to TransE~\citep{bordes2013translating}. We use the Euclidean norm for both scoring functions. We set the margin $\gamma = 1$ and train each model for 250 epochs without any hyperparameter tuning. 

We also experiment with varying the entity embedding dimension $\dim \Fc(s)$ across dimensions $[8,16,32,64]$ with dimensionality shared across entities. We use the same embedding dimensions for $\dim \Fc(r)$ and train models with all combinations where $\dim \Fc(r) <= \dim \Fc(s)$. We assume a single entity type for the schema of both datasets. 
Code for these experiments is available at \url{https://github.com/tgebhart/sheaf_kg}.

\begin{figure}
    \centering
    \includegraphics[width=1.19\columnwidth,keepaspectratio]{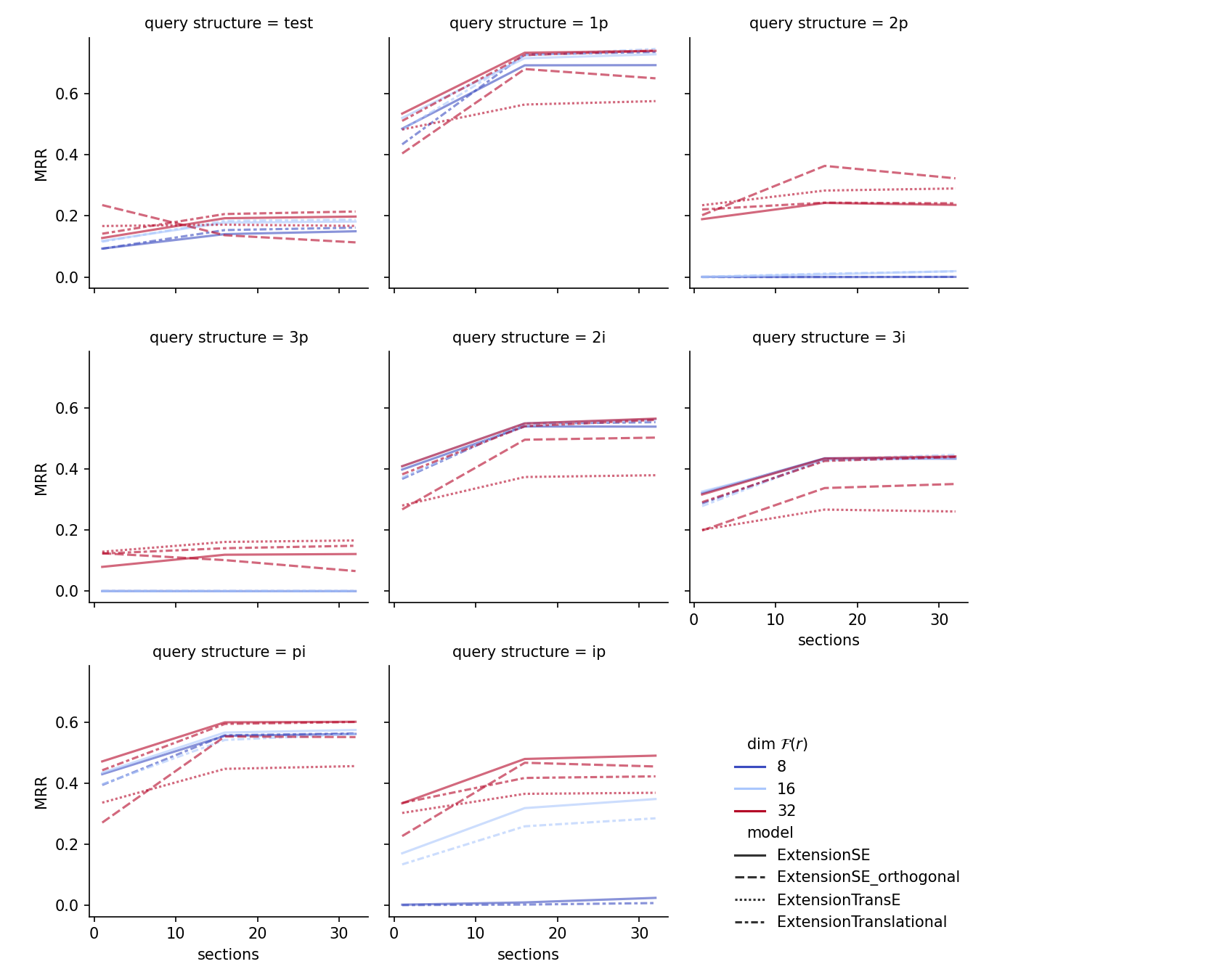}
    \caption{Triplet and complex query completion performance across varying number of sections and edge stalk dimensionality for each model class on NELL-995. Each sub-figure displays results for each complex query type and the test dataset. Line styles correspond to different model types and colors denote the dimensionality of the relation stalk space $\dim \Fc(r)$. The entity embedding dimension dim $\Fc(s)=32$ for each model.}
    \label{fig:sections_MRR_NELL}
\end{figure}

Figure~\ref{fig:sections_MRR_NELL} depicts a subset of these experimental results for NELL-995 (results for FB15k-237 in appendix) with entity embedding dimension fixed at 32 with varying number of (unregularized) sections and $\dim \Fc(r)$. Figure~\ref{fig:sections_MRR_NELL} indicates that square restriction maps ($\dim \Fc(r) = \dim \Fc(s)$) generally perform better than restriction maps which compress information across relations. The variants of the generalized Structured Embedding models generally achieve the highest performance on the complex query tasks, which is surprising given the competitive performance of translational models on the traditional completion tasks (test and 1p) within the broader literature. 
The higher performance of the \verb|ExtensionSE_orthogonal| model on path-structured queries compared to its performance on the test and 1p tasks highlights the differences in representational requirements for answering complex queries versus simple triplet queries, an observation reinforced by other works investigating compositional relational embeddings~\citep{guu2015traversing, sun2019rotate, tang2020orthogonal, cai2019group}.  

It is clear from Figure~\ref{fig:sections_MRR_NELL} that increasing the number of sections learned for each entity embedding improves overall performance across all model types without orthogonal parameterization of restriction maps. This result is not surprising, as increasing the number of sections increases model capacity. This interpretation is reinforced by Figure~\ref{fig:embedding_dimension_NELL} (appendix) which shows that increasing the entity embedding dimension $\dim \Fc(s)$ (an alternative route for increasing model capacity) also leads to increased performance.

We also compare the performance of the harmonic extension approach to a naive method for answering complex queries within the TransE model. This approach, as detailed in~\citet{guu2015traversing}, amounts to summation across all entity and relation embeddings involved in the complex query. Table~\ref{tab:transe_traversal_mrr} displays these results for models with embedding dimension 32 and 1 section. The similar performance between the two models on the path and intersectional queries is expected, as the harmonic extension over the identity restriction maps of the \verb|ExtensionTransE| model is very similar to the naive method for these simpler queries. However, on the more complex pi and ip queries, the harmonic extension approach significantly outperforms the naive approach.  

\input{tables/naive_vs_extension_easy_and_hard_mrr}

\section{DISCUSSION}\label{sec:discussion}

The sheaf-theoretic generalization of knowledge graph embedding presented in this paper provides a new perspective for integrating a number of prior embedding approaches within a cohesive theoretical framework. In addition to facilitating the comparison of these prior approaches within a common language, this generalization permits the incorporation of structured priors and expands the class of queries that can be answered using an embedding. By formalizing the relationship between typed knowledge graph schemas and their instantiation in knowledge graphs, this sheaf-theoretic framework provides a natural roadmap for the modeling of typed, hierarchical knowledge bases which provides further control over the representation of priors placed on the embeddings. By viewing knowledge graph embedding as sheaf learning, we have immediate access to tools for reasoning about the local and global consistency of embeddings through the sheaf Laplacian. Perhaps most importantly, this generalized perspective of knowledge graph embedding opens the door to an array of extensions like introducing uncertainty with a probabilistic approach, enforcing hierarchical typing within embeddings, or embedding knowledge graphs within more exotic categories.

This work is a preliminary exploration of the possibilities for using topology, sheaf theory, and spectral graph theory to understand knowledge graph embeddings. We anticipate that future work will deepen the theoretical foundations of this framework by exploring the representational capacity of sheaf embeddings with particular structures, and also refine the implementation and application to real datasets.

\bibliographystyle{abbrvnat}
\bibliography{main}

\include{supplement}

\end{document}

%% file: commands.tex
\usepackage{amsmath,amsfonts,amssymb,bm}
\usepackage{stmaryrd}

\renewcommand\vec{\bm}
\newcommand{\mat}[1]{{\bm{#1}}}

\newcommand{\Fc}{\mathcal{F}}

\DeclareMathOperator{\face}{\trianglelefteqslant}

\DeclareMathOperator*{\argmin}{arg\,min}

\DeclareMathAlphabet{\mathsfit}{\encodingdefault}{\sfdefault}{m}{sl}
\SetMathAlphabet{\mathsfit}{bold}{\encodingdefault}{\sfdefault}{bx}{n}

\newtheorem{definition}{Definition}

%% file: tables/naive_vs_extension_easy_and_hard_mrr.tex
\begin{table}[!ht]
    \centering
    \include{tables/easy/naive_vs_extension_mrr}
    \include{tables/hard/naive_vs_extension_mrr}
\caption{Comparative MRR performance of the extension and naive TransE approaches for ``easy'' (top) and ``hard'' (bottom) complex query answers on NELL-995 and FB15k-237 datasets.} 
\label{tab:transe_traversal_mrr}
\end{table}

%% file: tables/easy/naive_vs_extension_mrr.tex
\resizebox{\columnwidth}{!}{
\begin{tabular}{|l|l|l|l|l|l|l|l|}
\hline
    dataset & model & 2p & 3p & 2i & 3i & pi & ip \\ \hline
    NELL-995 & ExtensionTransE & \textbf{0.237} & \textbf{0.120} & 0.280 & 0.198 & \textbf{0.340} & \textbf{0.296} \\
    ~ & NaiveTransE & 0.164 & 0.082 & \textbf{0.281} & \textbf{0.205} & 0.271 & 0.104 \\ \hline
    FB15k-237 & ExtensionTransE & \textbf{0.084} & 0.050 & 0.157 & 0.147 & \textbf{0.119} & \textbf{0.109} \\ 
    ~ & NaiveTransE & 0.079 & \textbf{0.052} & \textbf{0.163} & \textbf{0.167} & 0.107 & 0.049 \\ \hline
\end{tabular}}

%% file: tables/hard/naive_vs_extension_mrr.tex
\resizebox{\columnwidth}{!}{
\begin{tabular}{|l|l|l|l|l|l|l|l|}
\hline
    dataset & model & 2p & 3p & 2i & 3i & pi & ip \\ \hline
    NELL-995 & ExtensionTransE & \textbf{0.060} & \textbf{0.054} & \textbf{0.165} & 0.234 & \textbf{0.126} & \textbf{0.079} \\ 
    ~ & NaiveTransE & 0.055 & 0.042 & 0.158 & \textbf{0.238} & 0.105 & 0.049 \\ \hline
    FB15k-237 & ExtensionTransE & 0.022 & \textbf{0.015} & \textbf{0.100} & \textbf{0.154} & \textbf{0.074} & \textbf{0.044} \\
    ~ & NaiveTransE & \textbf{0.025} & 0.013 & 0.093 & 0.153 & 0.064 & 0.024 \\ \hline
\end{tabular}}

%% file: supplement.tex

%
%





%

%

\onecolumn
\appendix
\section{OTHER EMBEDDING MODELS}

Many knowledge graph embedding approaches may be decomposed into a combination of multiplicative and additive interactions of entity and relation vectors. We show in this section that these additive components correspond to sheaf-theoretic coboundary operators across incident edges and discuss how different embedding approaches alter this coboundary operator. The models discussed in this section are a non-exhaustive subset of the total number of models in existence, but we make an attempt to cover most of the popular choices.

\textbf{Structured Embedding}. One of the earliest approaches to embedding knowledge graphs is Structured Embedding (SE) \cite{bordes2011learning}. Structured Embedding models entities $\vec{x}_{h} \in \mathbb{R}^d$ as $d$-dimensional vectors and relations as a pair of $(d \times d)$-dimensional matrices $(\mat{R}_r^h, \mat{R}_r^t)$. The scoring function between entities is then $f^{\text{SE}}(h,r,t) = \| \mat{R}_r^h \vec{x}_{h} - \mat{R}_r^t \vec{x}_{t} \|$. Setting $\mat{R}_r^h = \mat{\Fc}_{h \face r}$ and $\mat{R}_r^t = \mat{\Fc}_{t \face r}$, we see the scoring function computes precisely boundary function of the sheaf $f^{\text{SE}}(h, r, t) = \|\mat{\Fc}_{h \face r}\vec{x}_h - \mat{\Fc}_{t \face r} \vec{x}_t\|$. In other words, SE attempts to learn entity and relation embeddings that minimize the local discrepancy between adjacent entities along each relation. Therefore,
$$\sum\limits_{(h,r,t)}f^{\text{SE}}(h,r,t)^2 = \vec{x}^T \mat{L} \vec{x}$$
where $\mat{L}$ is the sheaf Laplacian formed from the matrices $\mat{R}^{\bullet}_r$, and $\vec{x} = (\vec{x})_{v \in V} \in C^0(G;\Fc^G)$.

\textbf{Unstructured Model}. The unstructured model~\citep{bordes2014semantic}, often used as a baseline model, is equivalent to Structured Embedding when $\mat{R}^h_r = \mat{R}^t_r = \mat{I}$, and therefore also fits within our modeling framework. 

\textbf{TransX}. A number of related embedding methods have been developed which seek to model relations as translations in a vector space which we refer to as the TransX class of embedding methods. These models seek to find embeddings of triples $(\vec{x}_h, \vec{r}_r, \vec{x}_t)$ such that $g(\vec{x}_h, \vec{r}_r) \approx \vec{x}_t$ where $g$ is a simple function representing a translation-like operation within the chosen embedding space. 

As discussed in the main text, TransE \citep{bordes2013translating} is an early translation-based model which aims to find embeddings that result in 
\begin{equation}\label{eq:transe_supplement}
f^{\text{TransE}}(h, r, t) = \| \vec{x}_h + \vec{r}_r - \vec{x}_t \|^2
\end{equation} being small when $(h,r,t)$ is true and large otherwise. Here, both the entity and relation embeddings are vectors in $\mathbb{R}^d$.

We can formulate this kind of translational scoring within our sheaf-theoretic framework by viewing the relation vector as a $\vec{r}_r$ as a $1$-cochain across edge $r$. More formally, we wish to learn some $1$-cochain $\vec{r} \in C^1(G;\Fc^G)$, representing a choice of vectors over each relation type in the knowledge graph, such that the discrepancy of entity embeddings $\vec{x}_h$ and $\vec{x}_t$ across each relation $r$ is approximately equal to $\vec{r}_r$:
\begin{equation}
f^{\text{TransX}}(h, r, t) = \|\mat{\Fc}_{h \face r}\vec{x}_h + \vec{r}_r - \mat{\Fc}_{t \face r} \vec{x}_t\|^2.
\end{equation} This is equivalent in form to TransR~\citep{lin2015learning} when both restriction maps are equivalent at the head and tail of $r$. Taking $\mat{\Fc}_{h \face r} = \mat{\Fc}_{t \face r} = \mat{I}$, our scoring function simplifies to exactly Equation~\ref{eq:transe_supplement} and is thus equivalent to TransE embedding. 

\textbf{TorusE and RotatE}. More recently, RotatE \citep{sun2019rotate} was introduced as a hybrid between ComplEx and the TransX approach. RotatE computes embeddings $\vec{x}_h, \vec{x}_t, \vec{r}_r \in \mathbb{C}^d$ and scores triplets translationally: 
\begin{equation}
    f^{\text{RotatE}}(\vec{x}_h, \vec{r}_r, \vec{x}_t) = \| \vec{x}_h \circ \vec{r}_r - \vec{x}_t \|
\end{equation} where $\circ$ is the Hadamard product. We can encode this scoring function through restriction maps as follows. Taking our edge and node stalk spaces to be in $\mathbb{C}^d$,  setting $\mat{\Fc}_{h \face r}$ to be the diagonal matrix with $\vec{r}_r$ on the diagonal\footnote{Equivalently, we can represent $\vec{r}_r$ as a diagonal matrix with $e^{i\vec{\phi}_r}$ on the diagonal where $\vec{\phi}_r$ is a vector of phases ranging from $0$ to $2\pi$.}, and setting $\mat{\Fc}_{t \face r} = \mat{I}$, we obtain an equivalent score for $f^{\text{RotatE}}$. The TorusE model \citep{ebisu2017toruse} is a special case of RotatE where the modulus of the embeddings are fixed.

Finally, Yang et al.~\citep{yang2020nage} propose a number of embedding methods which fit within this sheaf embedding framework as fixed restriction maps which introduce both inductive priors on the interactions between, and therefore the embeddings of, entities.

\subsection{Models Without Sheaf Structure}

The sheaf-theoretic framework presented in the main text does not generalize all knowledge graph embedding approaches that have been proposed in the literature. In general, any model with a bilinear interaction between entity embeddings and relations does not have a simple representation in terms of cellular sheaves. Models of this form include the Neural Tensor Network~\citep{socher2013reasoning}, ComplEx (equivalently, HolE)~\citep{trouillon2016complex, nickel2016holographic, hayashi2017equivalence}, Rescal/Bilinear~\citep{jenatton2012latent,nickel2011three}, and QuatE~\citep{zhang2019quaternion}. TransH~\citep{wang2014knowledge} also does not conform to our modeling framework, but does conform once entities are in their post-projection form. Investigating the extent to which these bilinear models may be incorporated into the sheaf embedding form is an interesting avenue for future work. 

Recently, a distinct lineage of knowledge graph embedding models have been proposed which represent a departure from translational/bilinear classification given above. Targeting the task of complex logical query answering, models like BetaE~\citep{ren2020beta} and Query2Box~\citep{ren2020query2box} look to embed the queries themselves within some representational space. It is currently unclear to the authors whether this family of models is amenable to a sheaf-theoretic representation. Casting these models within our sheaf embedding framework may require introducing sheaves valued in other categories than vector space stalks with linear restriction maps. The basics of such a generalization are discussed in the following section.

\section{WORKED EXAMPLES}
\subsection{Knowledge Graph}
To clarify definitions \ref{def:knowledge_database_schema} and \ref{def:knowledge_database}, we present a simple example of a knowledge graph $G$ and schema $\mathcal{Q}$. Here the schema has two types: Person and Film, and two relations: "friends" and "favorite movie." This is represented as a graph with two vertices and two edges, one of which is a self-loop for the vertex Person. The graph $G$ has three entities of type Person and three entities of type Film, with the relations indicated by edges. The graph morphism $k$ sends each person in $G$ to the vertex Person in $\mathcal{Q}$, and each film in $G$ to the vertex Film in $\mathcal{Q}$. It also sends the edges in $G$ to their similarly labeled edges in $\mathcal{Q}$. 

It may be helpful to note that any subgraph of $G$ is a valid knowledge graph for schema $\mathcal{Q}$. However, to add a node or edge to $G$, we must also specify which node or edge of $\mathcal{Q}$ it corresponds to. If we were to add an edge between Primer and Fargo in $G$, we would no longer be able to interpret $G$ as a knowledge graph with schema $\mathcal{Q}$, because there is no corresponding edge in $\mathcal{Q}$ from Film to itself.

\begin{figure}
    \centering
    \includegraphics[width=4in]{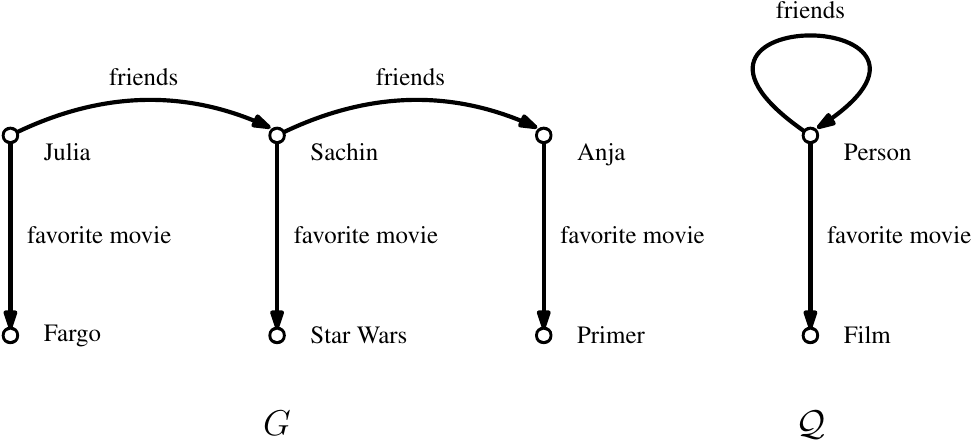}
    \caption{A simple knowledge graph $G$ with its schema $\mathcal{Q}$.}
    \label{fig:my_label}
\end{figure}

\subsection{Knowledge Sheaf}
Continuing this example, we can describe a knowledge sheaf $\Fc$ on $\mathcal{Q}$ and its pullback to $G$. For notational convenience, denote the node Person in $\mathcal{Q}$ by $u$ and the node Film by $v$, with the edge ``favorite movie'' denoted $e$ and the edge ``friends'' denoted $f$.
Let $\Fc(u) = \mathbb{R}^3$ and $\Fc(v) = \mathbb{R}^2$, with $\Fc(e) = \mathbb{R}^2$ and $\Fc(f) = \mathbb{R}$. We can now define the restriction maps of $\Fc$ using matrices:
\[
\begin{array}{cc}
    \Fc_{u\face e} = \begin{bmatrix}1 & 0 & 0 \\ 0 & 0 & 1\end{bmatrix} & \Fc_{v \face e} = \begin{bmatrix} 1 & 0 \\ 0 & 1 \end{bmatrix} \\
    \Fc_{u \face_h f} = \begin{bmatrix}0 & 1 & 0\end{bmatrix} & \Fc_{u \face_t f} =  \begin{bmatrix}0 & 1 & 0 \end{bmatrix} 
\end{array}
\]

Note that because the edge ``friends'' is a self-loop, we need to distinguish between the connection from $u$ to the head of $f$ ($u \face_h f$) and the connection from $u$ to the tail of $f$ ($u \face_t f$), as these may have two different restriction maps in general.

The pullback $\Fc^G$ of $\Fc$ to the knowledge graph $G$ would be unenlightening to describe completely, but we will give a few examples. For instance, because Julia is the head entity for an edge (call it $a$) of $G$ that maps to ``friends'' in $\mathcal{Q}$, $\Fc^G_{\text{Julia} \face a} = \Fc_{u \face_h f}$, and since Sachin is the tail entity for that edge, we get $\Fc^G_{\text{Sachin} \face a} = \Fc_{u \face_t f}$. Since Anja is the head entity for an edge $b$ that maps to ``favorite movie'' in $\mathcal{Q}$, we have $\Fc^G_{\text{Anja} \face b} = \Fc_{u \face e}$, and since Primer is the tail entity for $b$, we have $\Fc^G_{\text{Primer} \face b} = \Fc_{v \face e}$.

Choose the embeddings as follows:
\[
\begin{array}{ccc}
    x_{\text{Julia}} = \begin{bmatrix}1 \\ 1 \\ 0\end{bmatrix} & x_{\text{Sachin}} = \begin{bmatrix}0 \\ 1 \\ -1\end{bmatrix} & x_{\text{Anja}} = \begin{bmatrix}1 \\ 1 \\ 1\end{bmatrix} \\
    x_{\text{Fargo}} = \begin{bmatrix}1 \\ 0 \end{bmatrix} & x_{\text{Star Wars}} = \begin{bmatrix}0 \\ -1 \end{bmatrix} & x_{\text{Primer}} = \begin{bmatrix}1 \\ 1 \end{bmatrix} 
\end{array}
\]

The reader may check that this collection of data in fact defines a consistent sheaf embedding (Definition~\ref{def:consistent_sheaf_embedding}) of $G$. For instance, $x$ is consistent over the edge between Anja and Primer, because $\Fc_{u \face e} x_{\text{Anja}} = \begin{bmatrix}1 \\ 1\end{bmatrix} = \Fc_{v \face e} x_{\text{Primer}}$.

Symmetry of the relation ``friends'' is represented by the fact that $\Fc_{u \face_h f} = \Fc_{u \face_t f}$. Perhaps desirably in this instance, we see that this also forces the representation of ``friends'' to be transitive. The fact that each person can have only one favorite movie is represented by the fact that $\Fc_{v \face e}$ is a one-to-one linear map, so that for any possible Person embedding $x_u$, there is at most one Film embedding $x_v$ such that $\Fc_{v \face e} x_v = \Fc_{u \face e} x_u$.

\subsection{Complex Query}

Consider the problem of finding the maternal grandfather of a person in a knowledge database, from constituent relations ``is a child of'' and ``has gender.'' That is, $u_0$ is the person whose maternal grandfather we wish to find, and we seek entities $u_1$ and $u_2$ satisfying the following relations: $u_0$ is a child of $u_1$; $u_1$ is a child of $u_2$; $u_1$ has gender female; $u_2$ has gender male.

\begin{figure}
    \centering
    \includegraphics{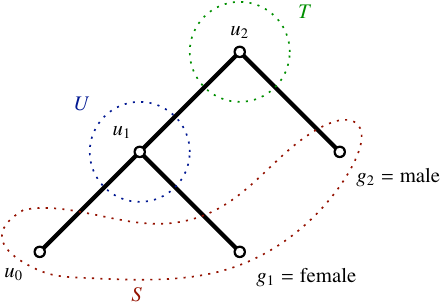}
    \caption{The template knowledge graph for finding the maternal grandfather $u_2$ of entity $u_0$. The interior $U$, source set $S$, and target set $T$ are labeled.}
    \label{fig:relationalgraph}
\end{figure}

There are five entities in this knowledge graph: the known source vertex $u_0$ of the desired relation, the unknowns $u_1$ and $u_2$, and the entities female and male. The boundary set $B$ consists of all vertices but $u_1$, and the source subset is $S = \{u_0, \text{female}, \text{male}\}$, while the target subset is simply $T = \{u_2\}$, as shown in Figure~\ref{fig:relationalgraph}. To find the maternal grandfather, we construct the sheaf on the relational graph $H$, find its Laplacian, and compute the Schur complement $\mat{L}_{\Fc^H}/\mat{L}_{\Fc^H}[U,U]$. Then we fix $\vec{y}_{u_0} = \vec{x}_{u_0}$, $\vec{y}_{g_1} = \vec{x}_{\text{female}}$ and $\vec{y}_{g_1} = \vec{x}_{\text{male}}$ to be the known embeddings of these entities, and search the entities for the entity $u_2$ whose embedding $\vec{x}_{u_2}$ gives the smallest value of $V(\vec{y}_B)$ when $\vec{y}_{u_2} = \vec{x}_{u_2}$. Note that by changing the values of $\vec{y}$ on the input set $S$, the computed Schur complement can also be used to find grandparents of any type for any initial entity $u_0$. We can thus think of the reduced matrix as describing a polyadic relation $R(u_0,u_2,g_1,g_2)$ which holds when $u_0$ is the child of someone of gender $g_1$ who is the child of $u_2$, who has gender $g_2$.

\section{HARMONIC EXTENSION FOR TRANSLATIONAL EMBEDDINGS}\label{sec:translational_harmonic_extension}
The problem of finding a harmonic extension in the affine or translational setting may be formulated as follows. Let $H$ be a graph, $\Fc$ a sheaf on $H$, and $B$ a subset of vertices of $H$ with complement $U$. We further assume that the translations are given by a $1$-cochain $\vec{b} \in C^1(H;\Fc)$, and we have a known boundary condition $\vec{x}_B$ defined on vertices in $B$. Harmonic extension is then the following optimization problem:

\[\min_{\vec{y} \in C^0(H;\Fc)} \|\mat{\delta} \vec{y} - \vec{b}\|^2 \quad \text{s.t. } \vec{y}_B = \vec{x}_B.\]
Expanding the objective gives an expression in terms of the Laplacian:
\[ \min_{\vec{y} \in C^0(H;\Fc)} \vec{y}^T\mat{L}\vec{y} - 2 \vec{b}^T\mat{\delta}\vec{y} + \vec{b}^T\vec{b} \quad \text{s.t. } \vec{y}_B = \vec{x}_B.\]
The Lagrange multiplier conditions for optimality are 
\begin{IEEEeqnarray*}{rCl}
    \mat{L}[U,U] \vec{y}_U + \mat{L}[U,B] \vec{y}_B &=& (\mat{\delta}^T \vec{b})_U\\
    \mat{L}[B,U] \vec{y}_U + \mat{L}[B,B] \vec{y}_B &=& \vec{\lambda} \\
    \vec{y}_B &=& \vec{x}_B.
\end{IEEEeqnarray*}
Since $\vec{\lambda}$ is free, these equations are readily solved for $\vec{y}_U$:
\begin{IEEEeqnarray*}{rCl}
\vec{y}_U &=& \mat{L}[U,U]^{-1}((\vec{\delta}^T \vec{b})_U - \mat{L}[U,B] \vec{x}_B)\\
        &=& -\mat{L}[U,U]^{-1}\mat{L}[U,B] \vec{x}_B + \mat{L}[U,U]^{-1}(\vec{\delta}^T \vec{b})_U \\
     &=& \vec{y}_U^\Fc + \mat{L}[U,U]^{-1}(\vec{\delta}^T \vec{b})_U,
\end{IEEEeqnarray*}
where $\vec{y}_U^\Fc$ is the harmonic extension of $\vec{x}_B$ for the sheaf $\Fc$ without the affine term $\vec{b}$. We now wish to compute the optimal value; this is 
\[\|\mat{\delta} \vec{y} - \vec{b}\|^2 = \vec{y}^T\mat{L}\vec{y} - 2 \vec{b}^T\mat{L}\vec{y} + \vec{b}^T\vec{b}.\]
We write $\vec{y} = \vec{y}_\Fc + \vec{y}_b$, where $\vec{y}_\Fc$ is the standard harmonic extension of $\vec{x}_B$ and $\vec{y}_b = \mat{L}[U,U]^{-1}(\vec{\delta}^T \vec{b})_U$ is the affine correction computed above (extended to $v \in B$ by zero). Then the optimal value is
\[\vec{y}_\Fc^T\mat{L} \vec{y}_\Fc + 2 \vec{y}_\Fc^T\mat{L}\vec{y}_b + \vec{y}_b^T\mat{L}\vec{y}_b - 2\vec{b}^T \mat{\delta} \vec{y}_\Fc - 2 \vec{b}^T \mat{\delta} \vec{x}_b + \vec{b}^T\vec{b}.\]  After substituting known values of $\vec{y}_\Fc$ and $\vec{y}_b$ in terms of $\vec{x}_B$ and dropping terms that do not depend on $\vec{x}_B$, we have
\[\vec{y}_\Fc^T \mat{L} \vec{y}_\Fc - 2 \vec{b}^T \mat{\delta} \vec{y}_\Fc.\]
This means that in order to calculate the affine harmonic extension cost, it suffices to compute the standard linear harmonic extension. 
The first term can be computed from $\vec{x}_B$ using the Schur complement 
$\mat{L}/\mat{L}[U,U]$, 
while the second term is equal to 
$2 \vec{b}^T \left(\mat{\delta}\vert_B - \mat{\delta}\vert_U\mat{L}[U,U]^{-1}\mat{L}[U,B] \right)\vec{x}_B$. 
This term is linear in $\vec{x}_B$ and hence is easily computed. 
    
Note that when $\vec{b} = 0$ this reduces to the standard harmonic extension problem, and hence gives a proof of the Schur complement formula given in the main text.

\section{HARMONIC EXTENSION AND MARGINALIZATION}

To better convey the role of harmonic extension in the complex query completion setting, it may be helpful to reframe this operation in a more familiar statistical language. Assume entity embeddings of a knowledge graph $G$ are distributed as $0$-mean multivariate normal: $p(\vec{x}_v) = \sqrt{(2\pi)^k \det \mat{\Sigma}}^{-1} \exp{-\frac{1}{2}(\vec{x}_v^T \mat{\Sigma}^{-1}\vec{x}_v)}$. For a set of boundary vertices $B$ and their complement $U$, their collection of embeddings $\vec{x}_H = (\vec{x}_B;\vec{x}_U)$ is also multivariate normal with zero mean and covariance $\mat{\Sigma}_H$ a block matrix with $\mat{\Sigma}_B$ and $\mat{\Sigma}_U$ as diagonal blocks and the covariance $\mat{\Sigma}_{BU}$ filling off-diagonal blocks. The conditional covariance of the boundary embeddings $\vec{x}_B$ given $\vec{x}_U$ is the Schur compelement of $\mat{\Sigma}_U$ in $\mat{\Sigma}_H$:
\begin{align*}
\mathbb{E}(\vec{x}_B \mid \vec{x}_U) &= \mathbb{E}(\vec{x}_B) + \mat{\Sigma}_{BU}\mat{\Sigma}^{-1}_U(\vec{x}_U - \mathbb{E}(\vec{x}_U)) = \mat{\Sigma}_{BU}\mat{\Sigma}_U^{-1}\vec{x}_U \\
\text{Cov}(\vec{x}_B \mid \vec{x}_U) &= \mat{\Sigma}_B - \mat{\Sigma}_{BU}\mat{\Sigma}^{-1}_U\mat{\Sigma}^T_{BU}.
\end{align*}
In this form, we see that the Laplacian of this knowledge sheaf $\mat{L}_{\Fc^G}$ corresponds to the inverse covariance matrix:
$$\text{Cov}(\vec{x}_B \mid \vec{x}_U)^{-1} = \mat{L}_{\Fc^G}[B,B] - \mat{L}_{\Fc^G}[B,U]\mat{L}_{\Fc^G}[U,U]^{-1}\mat{L}_{\Fc^G}[U,B].$$
It can be shown that the probability of observing any choice of embedding decomposes as a product of node and edge potential functions which are parameterized by $\mat{L}_{\Fc^G}$, providing a statistical interpretation of observing an entity embedding within a knowledge graph that is inversely proportional to the discrepancy it introduces with respect to its neighbors~\citep{malioutov2006walk}. This statistical interpretation is reminiscent of problems like covariance selection or graphic lasso, and the relationship between sheaf embedding, graphical likelihood estimation, and belief propagation may inspire future work. 

\section{SHEAVES AND CATEGORY THEORY}
While the algebraic properties of the theory of cellular sheaves rely on the linear structure of the vector spaces that serve as stalks, the theory may be developed in a more general setting. For this we adopt the language of category theory~(see \cite{riehl_category_2017,fong_seven_2018}). Let $G$ be a graph, which we view as a category with one object for each vertex and edge, and a unique morphism $v \face e: v \to e$ for each incident vertex-edge pair. To treat orientations properly, we also assume that morphisms are tagged with an orientation; that is, the morphism $v \face e$ also records whether the pair is to be viewed as defining the head or tail of the edge, which we will write as $v \face_h e$ or $v \face_t e$. This is particularly important for graphs with self-loops, where for a given vertex-edge pair $(v,e)$ there may be two distinct morphisms $v \face_h e$ and $v \face_t e$. (Formally, this means that the category representing $G$ is fibered over the category with two objects and two parallel non-identity morphisms.) 

A \emph{cellular sheaf} on $G$ valued in the data category $\mathcal C$ is a functor $\mathcal F: G \to \mathcal C$. We assume $\mathcal C$ is complete (i.e. has all small limits), and define the \emph{global sections} of $\Fc$ to be the limit $\lim \Fc$, an object of $\mathcal C$. The \emph{stalks} of $\Fc$ are the values the functor $\Fc$ takes on objects of $G$, and the \emph{restriction maps} are the values of $\Fc$ on the morphisms of $G$. Thus, for a vertex-edge pair $v \face e$, we have a restriction map $\Fc_{v \face e}: \Fc(v) \to \Fc(e)$, which is a morphism in $\mathcal C$. 

The pullback of $\mathcal C$-valued sheaves over a graph morphism $k$ is well-defined for graph morphisms that send edges to edges and vertices to vertices. It is constructed in the same way as for sheaves of vector spaces. For each vertex $v$, $k^*\Fc(v) = \Fc(k(v))$, and for each edge $e$, $k^*\Fc(e) = \Fc(k(e))$. Then $k^*\Fc_{v \face e} = \Fc_{k(v) \face k(e)}$. 

Since $\mathcal C$ is complete, we can define the object $C^0(G;\Fc) = \prod_{v} \Fc(v)$ in $\mathcal C$, the product of all vertex stalks of $\Fc$. The global sections of $\Fc$ naturally form a subobject of $C^0(G;\Fc)$; when $\mathcal C$ is a concrete category, we can think of sections of $\Fc$ as elements $(x_v)_{v\in V(G)} \in C^0(G;\Fc)$ such that for every edge $e = u \to v$, $\Fc_{u \face_h e} x_u = \Fc_{v \face_t e} x_v$. 

We can similarly define $C^1(G;\Fc) = \prod_{e \in E(G)} \Fc(e)$ as an object in $\mathcal C$. If $\mathcal C$ is the category of groups (or a subcategory thereof), we can define a coboundary map $\delta: C^0(G;\Fc) \to C^1(G;\Fc)$ by letting $(\delta x)_e = (\Fc_{u \face_h e} x_v)^{-1}(\Fc_{v \face_t e} x_v)$. When $\mathcal C = \mathbf{Vect}$, the category of vector spaces, this definition recovers the definition of the coboundary given in the paper.

We actually require slightly less structure to define a coboundary map; it is sufficient for $\mathcal C$ to be a category of group torsors. Recall that for a group $S$, an $S$-torsor $A$ is a set equipped with an action $\cdot: S \times A \to A$ and a division map $D: A \times A \to S$, such that $D(a,b) \cdot b = a$. We can formally think of $D(a,b)$ as being $ab^{-1}$, in which case the formula is the natural $ab^{-1} \cdot b = a$. The reason for this extension is to allow a broader class of maps (i.e. more than just group homomorphisms) between embedding spaces. A morphism of torsors is not required to preserve the origin, but does preserve the division operation. To compute the coboundary operator of a torsor-valued sheaf, which we think of as a function between the underlying sets of $C^0(G;\Fc)$ and $C^1(G;\Fc)$, we let $(\delta x)_e = D(\Fc_{u \face_h e} x_u,\Fc_{v \face_t e} x_v)$ for $e = u \to v$. The coboundary is then valued in a product of groups: if the stalk $\Fc(e)$ is an $S_e$-torsor, the coboundary $\delta x$ is in $\prod_{e} S_e$. When these groups are given a metric, we can then compute a cost function for a $0$-cochain $x$ by letting $U_\Fc(x) = \sum_{e \in E(G)} d((\delta_x)_e, 1_{S_e})$, where $1_{S_e}$ is the identity of the group $S_e$. The cost function $U_\Fc$ vanishes exactly on those $0$-cochains $x$ which are sections of $\Fc$.


Every vector space is an abelian group under addition, and the category of torsors over vector spaces is equivalent to the category $\mathbf{Aff}$ of vector spaces and affine maps. In this category, a morphism $f: V \to W$ is given by a formula of the form $f(\vec{v}) = T(\vec{v}) + \vec{b}$, where $T$ is a linear transformation $V \to W$ and $\vec{b} \in W$. The coboundary map of a sheaf valued in $\mathbf{Aff}$ is given on edges by $(\mat{\delta} \vec{x})_e = \mat{\Fc}_{v \face e} \vec{x}_v + \vec{b}_{v \face e} - \mat{\Fc}_{u \face e} \vec{x}_u - \vec{b}_{u \face e}$. This is equivalent to the coboundary map of a linear sheaf $\Fc$ with an edgewise affine correction term. Thus, for the purposes of knowledge graph embedding, working with sheaves valued in $\mathbf{Aff}$ is equivalent to using a sheaf valued in $\mathbf{Vect}$ and learning $\vec{x}$ and $\vec{b}$ such that $\mat{\delta} \vec{x} \approx \vec{b}$ rather than $\mat{\delta} \vec{x} \approx 0$. Passing to sheaves valued in $\mathbf{Aff}$ thus adds a translational component to the embedding model.

As a result, we can think of the relation between embeddings with group-valued sheaves and embeddings with torsor-valued sheaves as analogous the relationship between pure sheaf embeddings and sheaf embeddings with a translational component.


If we abandon the prospect of a translational component to the embedding, we can further relax our requirements on the data category $\mathcal C$. If $\mathcal C$ is a category of metric spaces, we can construct a measure of the discrepancy of a $0$-cochain $x \in C^0(G;\Fc)$ by 
\[V_{\Fc}(x) = \sum_{e \in E(G)} d_{\Fc(e)}(\Fc_{u \face e}(x_u),\Fc_{v \face e}(x_v)).\]
Optimizing this function with respect to the restriction maps $\Fc_{v \face e}$ and the 0-cochain $x$ produces a generalized knowledge graph embedding. In this setting, it is most reasonable to take $\mathcal C$ to be a category of Euclidean spaces and smooth (or at least almost-everywhere differentiable) maps, so that we can apply automatic differentiation and gradient descent.

The common thread in all these sheaf-theoretic constructions is the notion of comparison. To evaluate the plausibility of a relation holding between two entities, the entity embeddings are both transformed into a comparison space, and some measure of discrepancy between these transformed embeddings is calculated, giving a scoring function. Many commonly used knowledge graph scoring functions fit neatly into this framework, but some do not.

\section{ADDITIONAL EXPERIMENTAL DETAILS}

\begin{figure}[ht!]
    \centering
    \includegraphics[width=\columnwidth,height=\textheight,keepaspectratio]{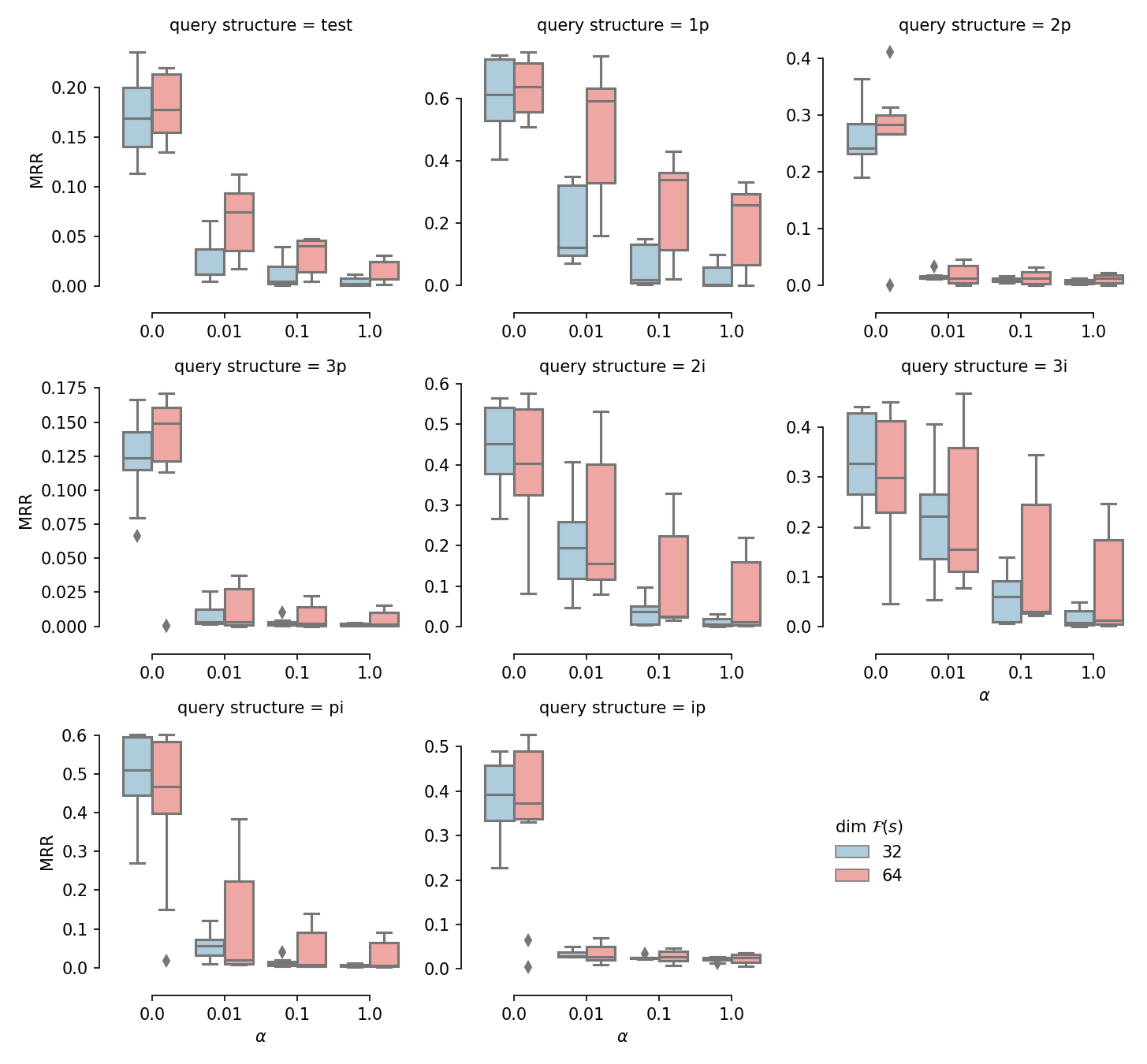}
    \caption{Effects of orthogonal regularization of sections on model performance by query structure and entity embedding dimensionality on NELL-995. Each color of box-and-whisker plot denotes the embedding and edge stalk dimensionality $\dim \Fc(s) = \dim \Fc(r)$. Results are aggregated across all model types. The x-axis denotes the orthogonal regularization weight $\alpha$. Evaluation on ``easy'' test queries.}
    \label{fig:reg_weight_MRR_by_query_structure}
\end{figure}

\begin{figure}[ht!]
    \centering
    \includegraphics[width=\columnwidth,height=\textheight,keepaspectratio]{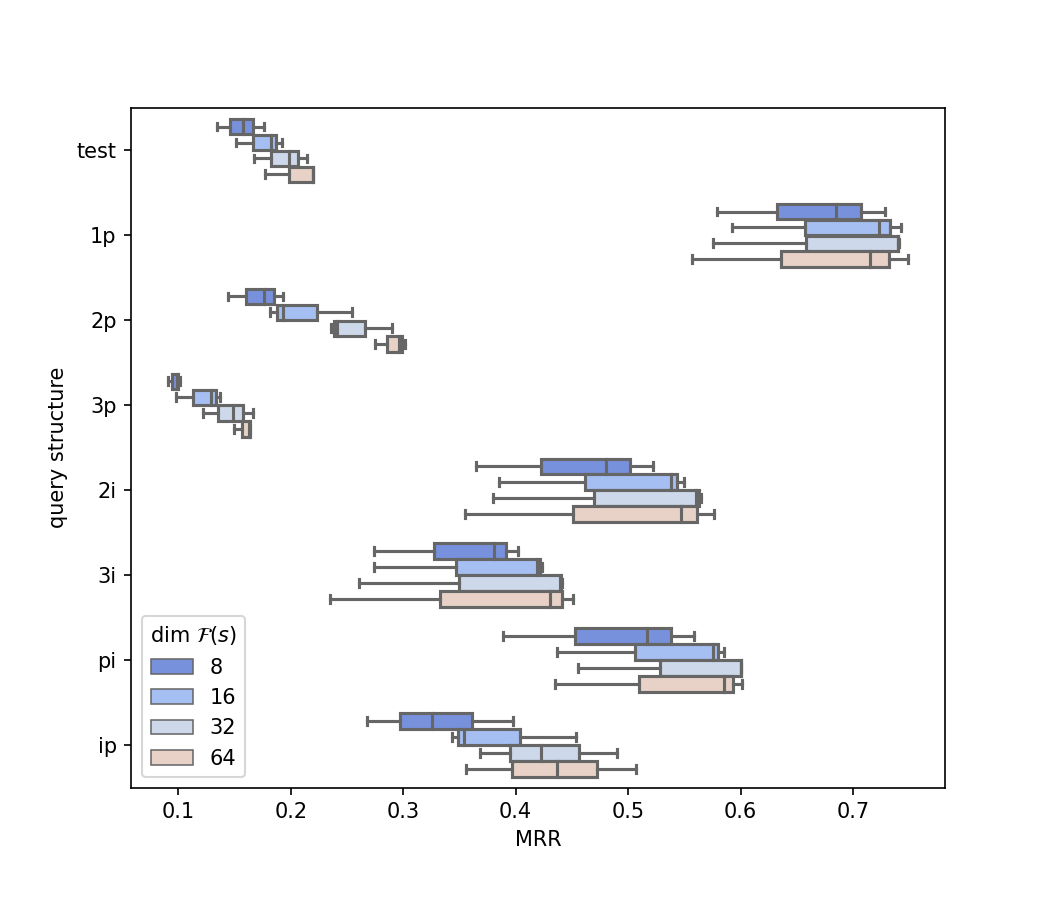}
    \caption{Performance on NELL-995 across varying embedding dimensions $\dim \Fc(s)$ for models with $\dim \Fc(s) = \dim \Fc(r)$ and 32 unregularized sections. Each box-and-whisker plot is aggregated across the three model types without orthogonal restriction map parameterization. Evaluation on ``easy'' test queries.}
    \label{fig:embedding_dimension_NELL}
\end{figure}

The ``easy'' test set for each complex query structure consists of query structures composed of individual triplets that have been seen at least once in the training set whereas the ``hard'' test set answers require at least one edge absent from the training set to be inferred~\citep{ren2020beta}. Because each complex query in the test set may have a number of feasible answers, we compute performance on the filtered dataset which amounts to reducing the ranking of entities by the number of acceptable answers for each query, computed independently for the ``easy'' and ``hard'' answer sets.

We compute the mean reciprocal rank (MRR) and Hits at 10 (Hits@10) from the rankings, according to their assigned score per test query, across all entities in the knowledge graph. Given a set of test triplets $\mathcal{T}_{\text{test}}$, MRR is defined as
$$\text{MRR} = \frac{1}{\vert\mathcal{T}_{\text{test}}\vert} \sum\limits_{(h,r,t) \in \mathcal{T}_{\text{test}}} \frac{1}{\text{rank}(t)}$$
For a ranking, denoted $\text{rank}(t)$, of the true entity $t$ score over the score of all other entities in the knowledge graph. The hits at $K$ metrics are defined as the proportion of true entities with ranking below some threshold $K$:
$$\text{Hits@}K = \frac{ \vert\{ (h,r,t) \in \mathcal{T}_{\text{test}} \mid \text{rank}(t) \leq K \}\vert }{\vert\mathcal{T}_{\text{test}}\vert}$$

We also trained versions of these models to learn multiple sections simultaneously, as detailed in Section~\ref{sec:learning_multiple_sections}. For such models, we were also interested in the performance effects of forcing these sections to being close to orthogonal, thereby increasing the size of the space of global sections. For these models, we vary the $\alpha$ orthogonal regularization parameter across $[0, 0.01, 0.1, 1]$ and vary the number of sections between $[1, 16, 32]$ to observe the effect of this regularization method. To investigate the performance effects of simultaneously learning multiple sections, we aggregated results across models with varying orthogonal regularization penalty values $\alpha$. The decrease in performance associated with orthogonal section regularization as depicted in Figure~\ref{fig:reg_weight_MRR_by_query_structure} implies that encouraging local orthogonality of the learned embedding cochains may be too strong a condition to produce useful embeddings. Instead, it appears that less-constrained representations are sufficient to solve the knowledge graph completion task, even in the context of complex logical queries. However, an interesting avenue of future work would be to investigate the extent to which this observation holds on complex queries whose structure has not been seen during training or on knowledge graphs with a greater diversity of abstract entity types which can appear within a variety of relational contexts.

\subsection{Baseline Comparison for Complex Queries}

\input{tables/both/vs_baselines_mrr}

We evaluated the $\verb|ExtensionSE|$ and $\verb|ExtensionTransE|$ models on the evaluation procedure used in \citet{ren2020beta}, allowing for direct comparison of performance between the harmonic extension models proposed in this work versus three state-of-the-art models for complex query reasoning: BetaE~\citep{ren2020beta}, Q2B~\cite{ren2020query2box}, and GQE~\citep{hamilton2018embedding}. 
The evaluation procedure amounts to evaluating on the ``hard'' test answer and filtering all ``easy'' answers for a given query in conjunction with the other correct ``hard'' answers. 
In short, we combine the ``easy'' and ``hard'' answer sets but only rank-score the ``hard'' answers. 

Table~\ref{tab:vs_baselines_mrr} displays the performance of these harmonic extension models versus the state-of-the-art results reported in \citet{ren2020beta}. 
Due to computational limitations, and to emphasize harmonic extension as a convenient drop-in method for extending traditional knowledge graph embedding models to conjunctive query reasoning, we set $\dim \mathcal{F}(s) = \dim \mathcal{F}(r) = 32$ and learn one section for each of the $\verb|ExtensionSE|$ and $\verb|ExtensionTransE|$ models. 
We also train these models using the traditional triplet scoring approach using margin ranking loss.
This is in contrast to the baseline models which embed the query itself and therefore require a training set composed of complex queries along with factual triplets. 

The model structure and training choices described above results in the $\verb|ExtensionSE|$ and $\verb|ExtensionTransE|$ models being structurally equivalent to the Structured Embedding and TransE models, respectively, with 32-dimensional embedding dimension. 
The number of tunable parameters for these models are on the order of $1$ million parameters for FB15k-237 and $2$ million parameters for NELL-995. 
This is in contrast to the baseline methods of~\citet{ren2020beta} which are on the order of $10$ million for FB15k-237 and $30-50$ million for NELL-995. 

Despite this difference in model size (90-95\% compression), the $\verb|ExtensionSE|$ and $\verb|ExtensionTransE|$ models are able to achieve performance within up to 50\% of these baselines on a number of conjunctive query structures.
These results imply the possibility that the harmonic extension approach to complex query reasoning may be able to achieve performance matching or exceeding that of BetaE and related query embedding models if we are able to find a base model--like Structured Embedding, TransE, or other embedding methods which fall within the sheaf-theoretic framework--that achieves 1p MRR performance in line with these baselines models. 
Such performance may be achievable, as previous works have shown that, for proper hyperparameter choices, models like TransE or RotatE can achieve MRR performance above 0.3 on FB15k-237, closing the gap between the baselines and the extension models trained in this work. 
We leave proper experimental validation of this observation to future work.  

\begin{figure}[ht!]
    \centering
    \includegraphics[width=1.2\textwidth,height=\textheight,keepaspectratio]{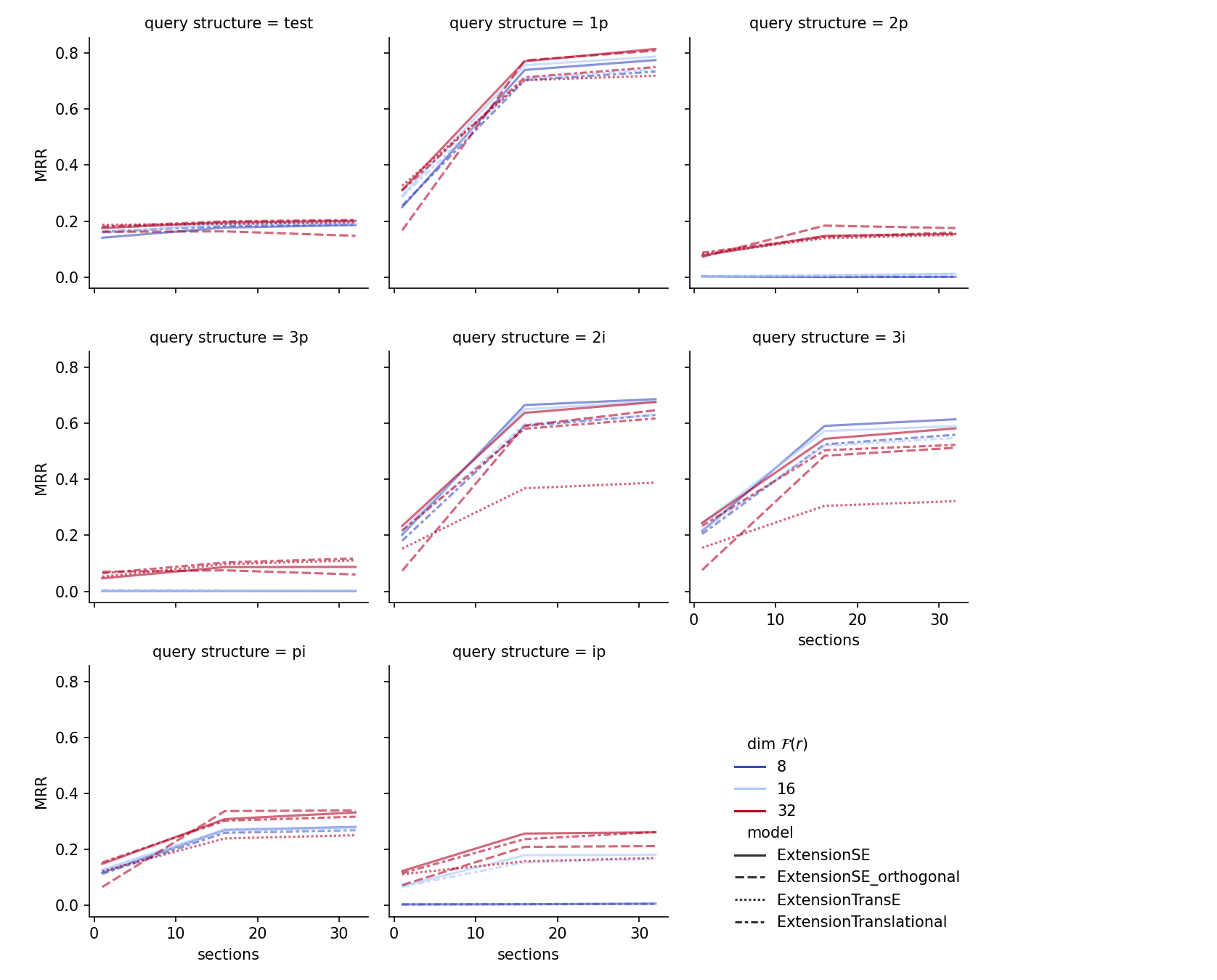}
    \caption{Triplet and complex query completion performance across varying number of sections and edge stalk dimensionality for each model class on FB15k-237. Each sub-figure displays the results for each complex query type, including the test dataset. Line styles correspond to different model types and their colors denote the dimensionality of the relation stalk space $\dim \Fc(r)$. The entity embedding dimension dim $\Fc(s)$ is held at $32$ for each model. Evaluation on ``easy'' test queries.}
    \label{fig:sections_MRR_FB15k-237}
\end{figure}

\begin{figure}[ht!]
    \centering
    \includegraphics[width=\textwidth,height=\textheight,keepaspectratio]{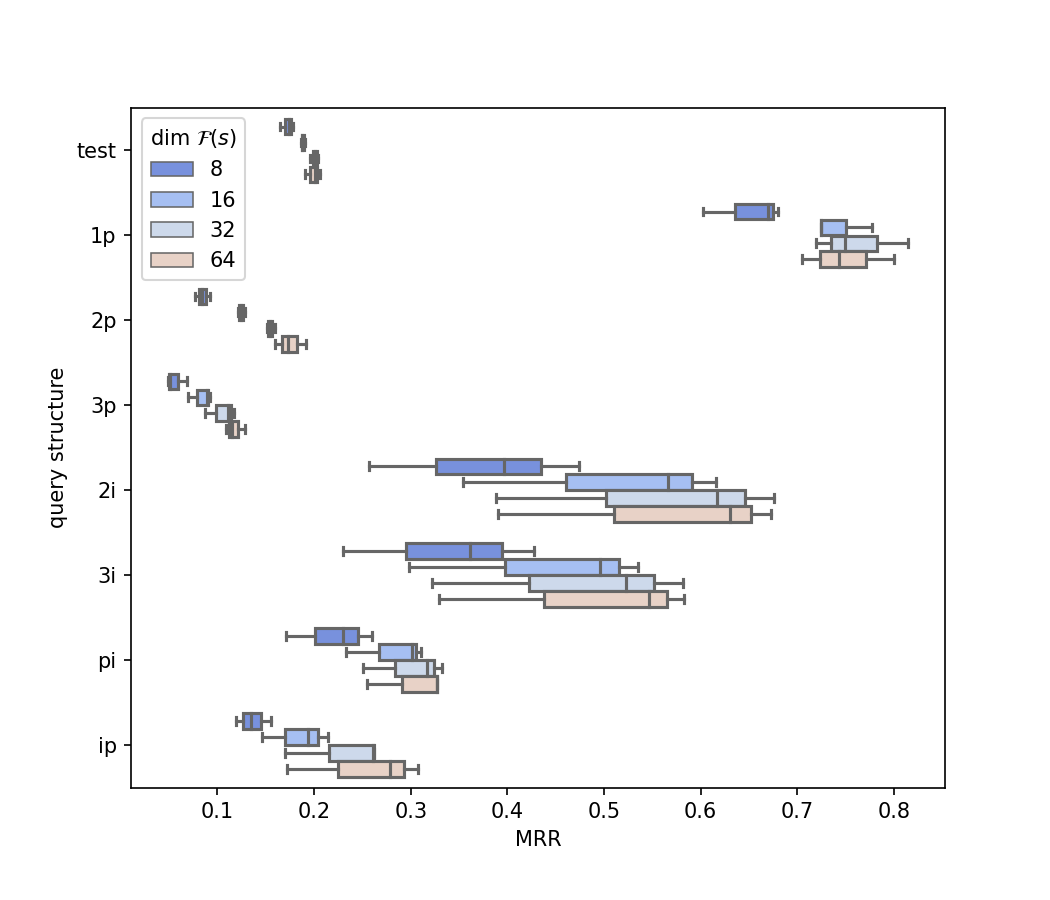}
    \caption{Performance on FB15k-237 across varying embedding dimensions $\dim \Fc(s)$ for models with $\dim \Fc(s) = \dim \Fc(r)$ and 32 unregularized sections. Each box-and-whisker plot is aggregated across the three model types without orthogonal restriction map parameterization. Evaluation on ``easy'' test queries}
    \label{fig:embedding_dimension_FB15k-237}
\end{figure}

\begin{figure}[ht!]
    \centering
    \includegraphics[width=\textwidth,height=\textheight,keepaspectratio]{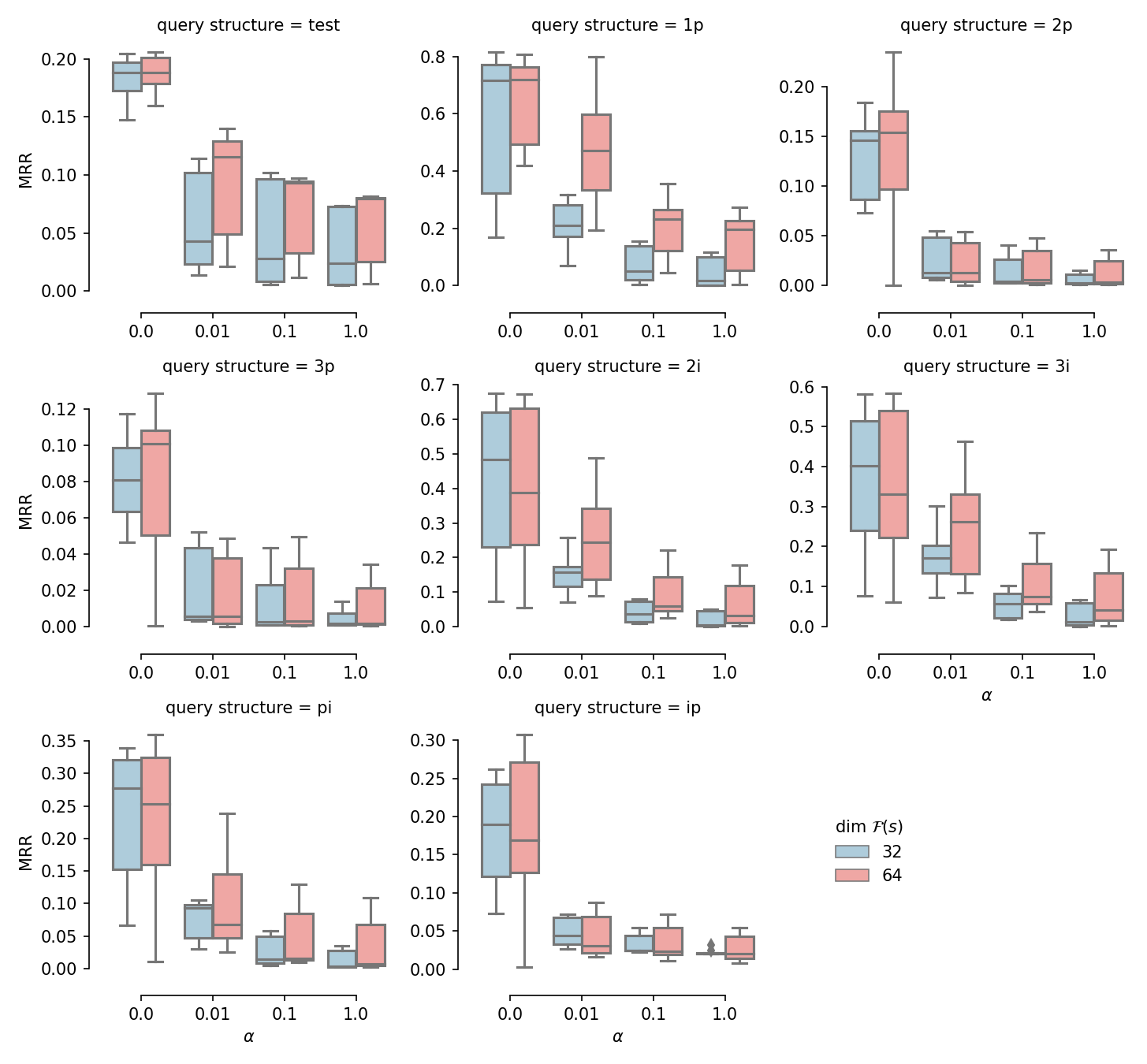}
    \caption{Effects of orthogonal regularization of sections on model performance by query structure and entity embedding dimensionality on FB15k-237. Each color of box-and-whisker plot denotes the embedding and edge stalk dimensionality $\dim \Fc(s) = \dim \Fc(r)$. Results are aggregated across all model types. The x-axis denotes the orthogonal regularization weight $\alpha$. Evaluation on ``easy'' test queries.}
    \label{fig:reg_weight_FB15k-237_by_query_structure}
\end{figure}

\input{tables/naive_vs_extension_easy_and_hard_hits_at_10}

\vfill


%% file: tables/both/vs_baselines_mrr.tex
\begin{table}[!ht]
    \centering
    \begin{tabular}{|l|l|l|l|l|l|l|l|l|l|}
    \hline
        dataset & model & 1p & 2p & 3p & 2i & 3i & pi & ip & parameters (M) \\ \hline
        FB15k-237 & BetaE & 0.390 & 0.109 & 0.100 & 0.288 & 0.425 & 0.224 & 0.126 & 14.3 \\ 
        ~ & Q2B & 0.406 & 0.094 & 0.068 & 0.295 & 0.423 & 0.212 & 0.126 & 6.8 \\ 
        ~ & GQE & 0.350 & 0.072 & 0.053 & 0.233 & 0.346 & 0.165 & 0.107 & 13.3 \\ \cline{2-10}
        ~ & ExtensionSE & 0.197 & 0.024 & 0.013 & 0.085 & 0.110 & 0.055 & 0.040 & 1.4 \\ 
        ~ & ExtensionTransE & 0.207 & 0.027 & 0.021 & 0.071 & 0.103 & 0.045 & 0.047 & 0.5 \\ \hline
        NELL & BetaE & 0.530 & 0.130 & 0.114 & 0.376 & 0.475 & 0.241 & 0.143 & 53.4 \\
        ~ & Q2B & 0.422 & 0.140 & 0.112 & 0.333 & 0.445 & 0.224 & 0.168 & 26.3 \\ 
        ~ & GQE & 0.328 & 0.119 & 0.096 & 0.275 & 0.352 & 0.184 & 0.144 & 52.3 \\ \cline{2-10}
        ~ & ExtensionSE & 0.159 & 0.037 & 0.024 & 0.109 & 0.118 & 0.125 & 0.072 & 2.8 \\ 
        ~ & ExtensionTransE & 0.187 & 0.035 & 0.028 & 0.083 & 0.079 & 0.088 & 0.069 & 2.0 \\ \hline
    \end{tabular}
\caption{Performance of logical query baselines versus the harmonic extension solution for simple StructuredEmbedding and TransE models using the evaluation setup of \citet{ren2020beta}. The first seven columns measure MRR performance for each complex query structure while the final column measures the number of trainable parameters (in millions). The Extension models set $\dim \mathcal{F}(s) = \dim \mathcal{F}(r) = 32$ and contain one section. Performance metrics for BetaE, Q2B, and GQE models taken from \citet{ren2020beta}.}
\label{tab:vs_baselines_mrr}
\end{table}

%% file: tables/naive_vs_extension_easy_and_hard_hits_at_10.tex
\begin{table}[!ht]
    \centering
    \input{tables/easy/naive_vs_extension_hits_at_10} \\
    \vspace{1em}
    \input{tables/hard/naive_vs_extension_hits_at_10}
\caption{Comparative H@10 performance of the extension and naive TransE approaches for ``easy'' (top) and ``hard'' (bottom) complex query answers on NELL and FB15k-237 datasets.} 
\label{tab:transe_traversal_easy}
\end{table}

%% file: tables/easy/naive_vs_extension_hits_at_10.tex
\begin{tabular}{|l|l|l|l|l|l|l|l|}
    \hline
        dataset & model & 2p & 3p & 2i & 3i & pi & ip \\ \hline
        NELL-995 & ExtensionTransE & 0.355 & 0.201 & 0.467 & 0.368 & 0.499 & 0.427 \\
        ~ & NaiveTransE & 0.280 & 0.148 & 0.488 & 0.403 & 0.451 & 0.216 \\ \hline
        FB15k-237 & ExtensionTransE & 0.163 & 0.107 & 0.343 & 0.341 & 0.253 & 0.207 \\
        ~ & NaiveTransE & 0.162 & 0.095 & 0.362 & 0.363 & 0.229 & 0.097 \\ \hline
\end{tabular}

%% file: tables/hard/naive_vs_extension_hits_at_10.tex
\begin{tabular}{|l|l|l|l|l|l|l|l|}
\hline
    dataset & model & 2p & 3p & 2i & 3i & pi & ip \\ \hline
    NELL-995 & ExtensionTransE & 0.101 & 0.087 & 0.298 & 0.393 & 0.198 & 0.124 \\ 
    ~ & NaiveTransE & 0.086 & 0.060 & 0.308 & 0.402 & 0.173 & 0.088 \\ \hline
    FB15k-237 & ExtensionTransE & 0.049 & 0.028 & 0.213 & 0.312 & 0.158 & 0.086 \\ 
    ~ & NaiveTransE & 0.049 & 0.024 & 0.213 & 0.307 & 0.142 & 0.045 \\ \hline
\end{tabular}